\documentclass[twoside,a4,12p]{report} %,draft,openright]

\usepackage{amsmath}
\usepackage{mathtools}
\usepackage{epsf,graphicx}
\usepackage{caption}
\usepackage{subcaption}
\usepackage{leftidx}

\usepackage{graphicx}
\usepackage{subfig}
\usepackage{multirow}
\usepackage{array}
\usepackage{url}
\usepackage{float}

\usepackage{latexsym,amssymb}
\usepackage{setspace,cite}
\usepackage{anysize}
\marginsize{4cm}{2.5cm}{4cm}{4cm}

\usepackage{fancyhdr}

\begin{document}
\newpage

%Puts page numbering of preamble in roman and of main body of thesis in
%arabic. Also defines how chapters and sections are made
\pagenumbering{arabic}
\setcounter{page}{1} \pagestyle{fancy}
\renewcommand{\chaptermark}[1]{\markboth{\chaptername%
\ \thechapter:\,\ #1}{}}
\renewcommand{\sectionmark}[1]{\markright{\thesection\,\ #1}}

%DEFINES TITLE PAGE, and contains abstract, acknowledgements, etc.

%%%%%%%%%%%%%%%%%%%%%%%%%%%%%%%%%%%%%%%%%%%%%%%%%%%%%%%%%%%%%%%%%%%%%%%%%%%
% This is a sample header for a sample dissertation. Fill in the name,
% and the other information. LaTeX will work out the table of
% content, the list of figures and of tables for you.
%%%%%%%%%%%%%%%%%%%%%%%%%%%%%%%%%%%%%%%%%%%%%%%%%%%%%%%%%%%%%%%%%%%%%%%%%%%

\newpage
\thispagestyle{empty}

% ******* Title page *******
% **************************
\vspace*{-6cm}

\begin{center}
	\includegraphics{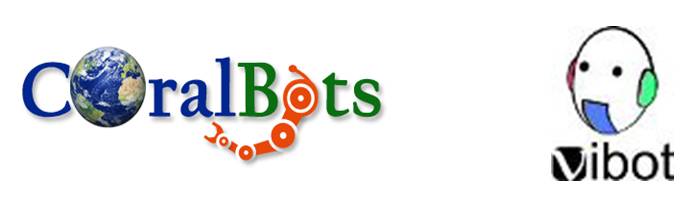}
\end{center}

\vspace{1.5cm}

\begin{center}
%{\Huge Machine Vision for Coralbots\\} 
{\Huge Sparse Coral Classification Using Deep Convolutional Neural Networks\\} 
\vspace{1cm} 
{\large
\textbf{Mohamed Elsayed Elawady}\\
\vspace{0.3cm}
Supervised by\\
\vspace{0.1cm}
Dr. Neil Robertson \\
Prof. David Lane\\
\vspace{1.5cm}
Centre Universitaire Condorcet\\
University of Burgundy\\
\vspace{0.5cm}
Department of Computer Architecture and Technology\\
University of Girona\\
\vspace{0.5cm}
School of Engineering and Physical Sciences\\
Heriot-Watt University\\
}
\end{center}

\vspace{0.6cm}
\begin{center}
{\large A Thesis Submitted for the Degree of \\MSc Erasmus Mundus
in Vision and Robotics (VIBOT) \\\vspace{0.3cm} $\cdot$ 2014
$\cdot$}
\end{center}
\singlespacing

\begin{center}
	\includegraphics{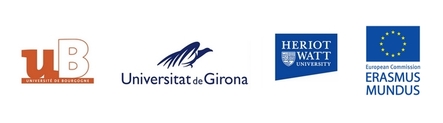}
\end{center}

%ABSTRACT
\begin{abstract}

Autonomous repair of deep-sea coral reefs is a recent proposed idea to support the oceans’ ecosystem in which is vital for commercial fishing, tourism and other species. This idea can be operated through using many small autonomous underwater vehicles (AUVs) and swarm intelligence techniques to locate and replace chunks of coral which have been broken off, thus enabling re-growth and maintaining the habitat.

The aim of this project is developing machine vision algorithms to enable an underwater robot to locate a coral reef and a chunk of coral on the seabed and prompt the robot to pick it up. Although there is no literature on this particular problem, related work on fish counting may give some insight into the problem. The technical challenges are principally due to the potential lack of clarity of the water and platform stabilization as well as spurious artifacts (rocks, fish, and crabs).

We present an efficient sparse classification for coral species using supervised deep learning method called Convolutional Neural Networks (CNNs). We compute Weber Local Descriptor (WLD), Phase Congruency (PC), and Zero Component Analysis (ZCA) Whitening to extract shape and texture feature descriptors, which are employed to be supplementary channels (feature-based maps) besides basic spatial color channels (spatial-based maps) of coral input image, we also experiment state-of-art preprocessing underwater algorithms for image enhancement and color normalization and color conversion adjustment.

Our proposed coral classification method is developed under MATLAB platform, and evaluated by two different coral datasets (University of California San Diego's Moorea Labeled Corals, and Heriot-Watt University's Atlantic Deep Sea).
\vspace*{5cm}

\begin{center}
\begin{quote}
\it We are part of a living world, not apart from it
\end{quote}
\end{center}
\hfill{\small Sylvia Earle}

\end{abstract}

\doublespacing

\pagenumbering{roman}
\setcounter{page}{1} \pagestyle{plain}

\tableofcontents

\listoffigures
\listoftables

\chapter*{Acknowledgments}
\addcontentsline{toc}{chapter}
         {\protect\numberline{Acknowledgments\hspace{-96pt}}}

I'd like to acknowledge Computer Vision Group at University California San Diego for the use of Moorea Labeled Corals dataset. I'd like also to acknowledge Rasmus Berg Palm at Technical University of Denmark for publishing MATLAB code for Deep Learning Toolbox, Oscar Beijbom for implementation of point-based coral classification, and Stephane Bazeille for implementation of underwater image preprocessing.

I'd like to thank Neil Robertson and David Lane for giving me an amazing masters thesis opportunity to work on a Heriot-Watt crucible project (Coralbots) with an interdisciplinary team of researchers from Life Sciences, Computer Science and Engineering. I'd like also to thank cold-water coral group (Lea-Anne Henry, Murray Roberts, and Laurence De Clippele) for providing different types of proficient information about coral reefs and maintaining new cold-coral dataset.

I can't find enough words expressing my gratitude to my family, you are the reason of every success i have done so far.  Many thanks to VIBOT classmates who share their cultures and background experiences with me, VIBOT family (David Fofi, Herma Adema-Labille, and David Arnoud), and all members of VisionLab Team (especially Rick, Rolf, Puneet, and Roushanak) for providing me ideas and suggestions through my masters research. 

Finally, I like to thank European Commission for granting Erasmus Mundus Scholarships which gave me an opportunity to study in three different universities (University of Burgundy, University of Girona, Heriot-Watt University) at three different countries (France, Spain and Scotland).

\pagestyle{fancy}

\newpage

%sets up headers for lefthand and righthand pages. To alter, edit
%these lines and the chaptermark/sectionmark lines above
\addtolength{\headheight}{3pt} \fancyhead{}
\fancyhead[LE]{\sl\leftmark} \fancyhead[LO,RE]{\rm\thepage}
\fancyhead[RO]{\sl\rightmark} \fancyfoot[C,L,E]{}
\pagenumbering{arabic}

%\singlespacing
%\doublespacing
\onehalfspacing
\chapter{Introduction} \label{chap:intro}
This chapter presents a basic background about coral species, thesis contribution to classify coral images through deep learning methods, and finally thesis outline for next chapters.

\section{Background} \label{sect:bkgrnd}
\begin{figure}
	\centering
		\includegraphics[width=0.9\textwidth]{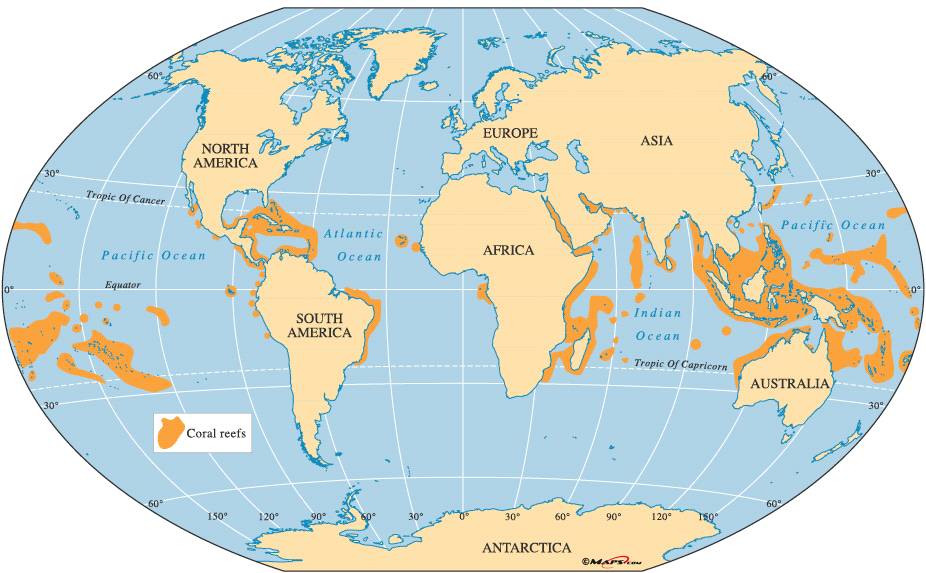}
		\captionsetup{justification=centering,margin=2cm}
	\caption{Distribution of Coral Reefs around the world (Copyright Friedrich Von Steuben Metropolitan Science Center)}
	\label{fig:Coral_Distro}
\end{figure}

\begin{figure}
	\centering
		\includegraphics[width=0.9\textwidth]{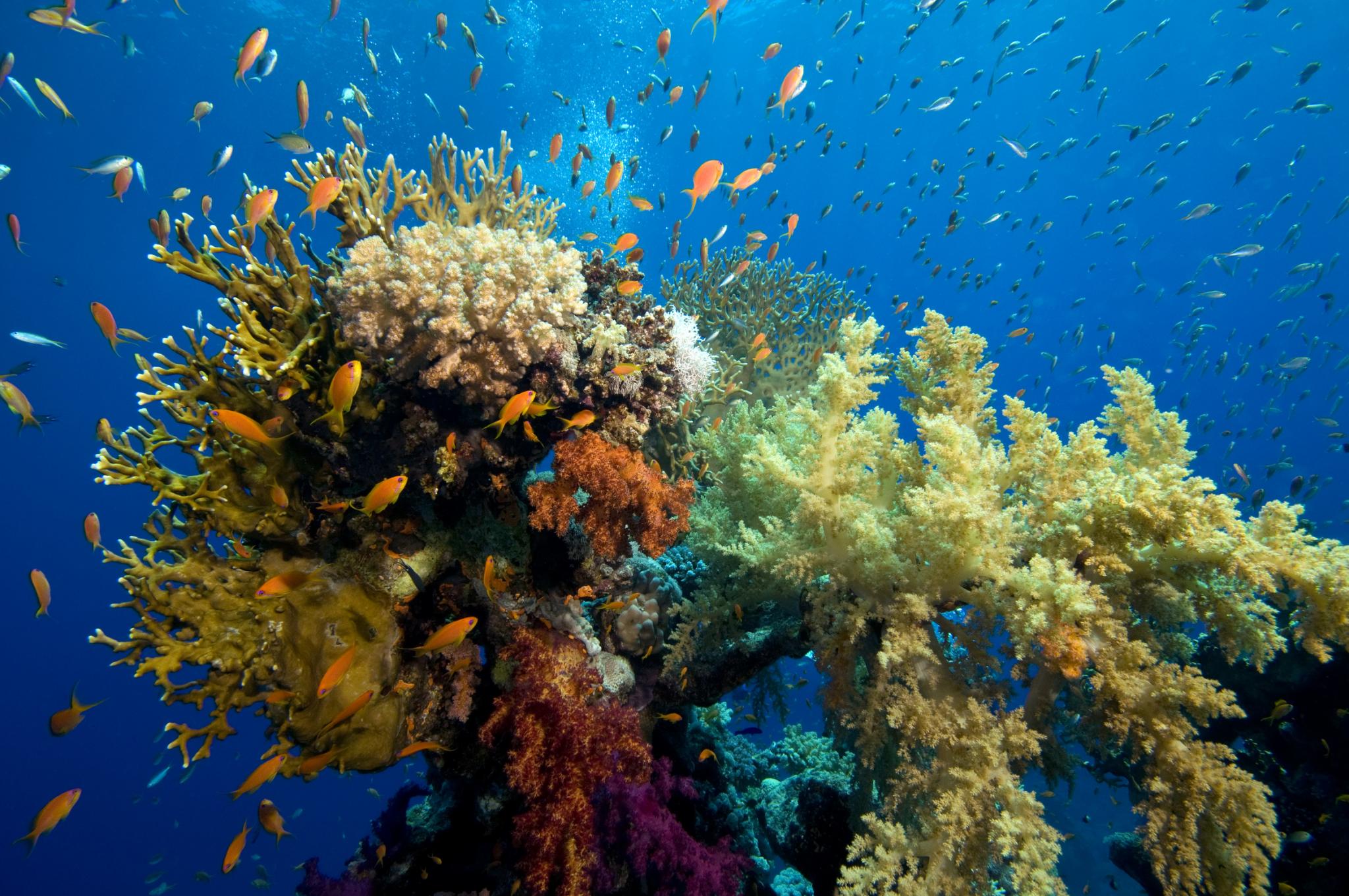}
	\caption{Great Barrier Reef, Australia (Copyright Zicasso)}
	\label{fig:Aus_GBR}
\end{figure}

Coral reefs are living organisms consisting of very small animals called ``polyps"', and exist in more than 200 countries' aquatic spaces (seas and oceans) where sea temperatures range between 25 and 290 Celsius (as shown in figure~\ref{fig:Coral_Distro}). They provide a main livelihood source for around 30 million people through (food in fisheries, income in tourism and materials in pharmaceuticals, and coastal protection of habitation and farmland in erosion and storms). World's largest coral reef system is Great Barrier Reef \cite{Rep2006} in the north-east coast of Australia (please see figure~\ref{fig:Aus_GBR}) spreading on thousands of coastal kilometers and hundreds of islands, and it contains around three thousands individual coral reefs \cite{De2012} and represents world's biggest single structure made by living organisms which can be seen from outer-space.

Hundreds of different species of corals exist around the world, in which are generally classified into hard and soft corals \cite{ICRANOnline}. Hard corals grow in colonies to build huge reef blocks. Seawater provide calcium to corals in order to build their skeletons. The living corals represents a very small part of the overall reef structure. Soft corals construct plants or trees and do not have stony skeletons. Soft corals can be found in different aquatic regions (tropical shallow sea and cold deep sea).

As the number of images and databases continues to rapidly increase at many environmental research centers and aquatic-based agencies (i.e. International Coral Reef Action Network, and National Oceanic \& Atmospheric Administration) due to latest technologies in image acquisition using different autonomous underwater vehicles. Coral ecologists and environmental scientists already collected millions of coral images and thousands of hours of underwater videos, and they need massive number of hours to annotate every pixel inside each coral image or video frame such that this full manual segmentation will be time consuming and increase the ratio between labeled and unlabeled images across time. Uniform random point sampling is a sufficient solution for coral research statistics using image annotation software (i.e. Coral Point Count \cite{Kohler2006} by National Coral Reef Institute). In this software,  images are manually annotated through coral experts by selecting some random pixels (10-200) in the target image, and classifying those pixels respect to predefined coral classes. A typical survey states \cite{Beijbom2012} that more than 400 hours are required to annotate 1000 images (around 200,000 coral labeled points). Automated image analysis is rapidly emerging as a promising cost-effective tool to annotate images for e.g., coral cover, health and species composition in both shallow and deep coral reef settings.

\section{Contribution}
An efficient sparse classification for coral species is introduced using most recent machine learning technique ``Deep Learning" which is a set of algorithms \cite{Bengio2013} that attempts to model high-level abstractions in data by using architectures composed of multiple non-linear transformations. Toronto's Hinton \cite{Hinton2006}, Montreal's Bengio \cite{Bengio2007}, and New York's LeCun \cite{Poultney2006} pioneer deep learning idea to be a new generation of artificial neural networks to allow machines to learn recognizing patterns in everything from audio/visual data \cite{Lee2009a,Lee2009,Wang2010} to spoken language \cite{Collobert2008} to handwriting \cite{Salakhutdinov2012}. \\
Two most-popular algorithms are Convolutional Neural Network (CNN) and Deep Belief Net (DBN). first is a special kind of multi-layer feed-forward supervised neural network which is designed to recognize visual objects directly from spatial-based images with minimal or without preprocessing, it consists of three layers: (1) feature extraction (convolution layer), (2) shift and distortion invariance (sub-sampling layer), (3) classification (output layer). later \cite{Deng2014} is an unsupervised probabilistic generative models composed of multiple layers of stochastic, hidden variables. The top layers have undirected, symmetric connections between them and the lower layer receives directed connections from the layer above.

\section{Thesis Outline}
Thesis is divided in 5 chapters: Introduction, Problem Definition, State of the Art, Methodology, Results and Conclusions. In chapter 2, it presents the coral threats and survival solutions. In chapter 3, it discusses the related research in coral detection, and introduces recent supervised machine learning algorithm in deep learning (convolutional neural networks) \& its applications in object classification and recognition. In chapter 4, it shows overview of the proposed coral classification method and explains in details each phase. In chapter 5, it evaluates the proposed method's results in qualitative and quantitative way. Finally in chapter 6, it summarizes conclusions, method's limitations, and future work.
\chapter{Problem Definition} \label{chap:problem}
This chapter presents importance of coral reefs, its main threads due to human and environmental effects, finally manual and automatic transplantation process as survival coral solutions.  

\section{Introduction}
Coral reef ecosystems provide for over half a million people: they create substantial socioeconomic benefits from tourism and fisheries while providing coastal protection, enhancing biodiversity and contributing to carbon sequestration that mitigates global warming \cite{Foley2010, Compilation2008}. Global conservation of reefs and their resources in a world characterized by multiple stressors and disturbances will require unified efforts to create international marine and climate policies alongside local adaptive community management tools \cite{Kennedy2013}. However these policies and tools must also be cost-effective and promote public and stakeholder stewardship of coral reefs \cite{Ahmed2004}.

\section{Main Threats}
\begin{figure}
	\centering
		\includegraphics[width=1.00\textwidth]{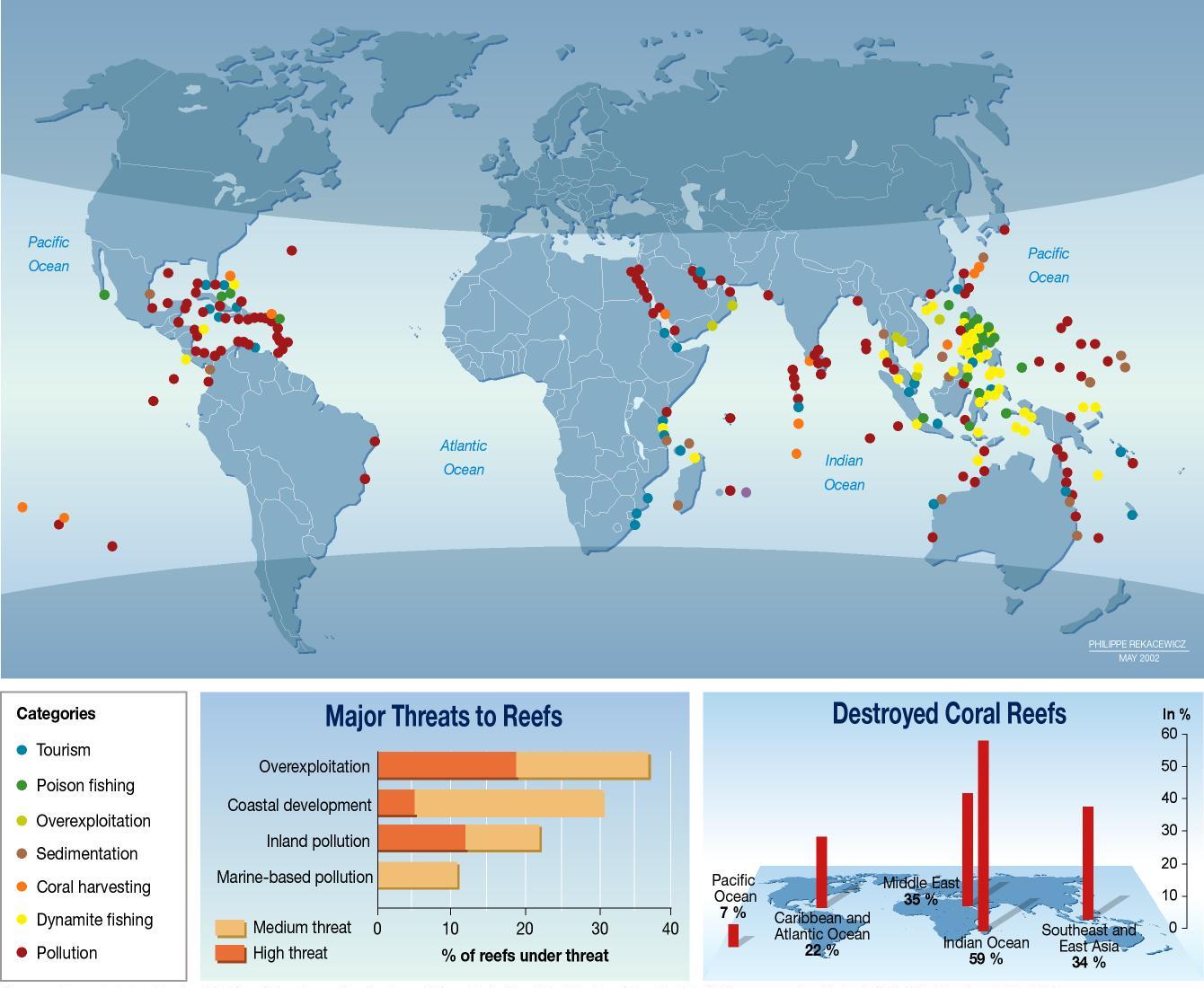}
	\caption{Threads to Coral Reefs \cite{Diop2002}}
	\label{fig:0302-coralreefs-EN}
\end{figure}

\begin{figure}
	\centering
		\includegraphics[width=1.00\textwidth]{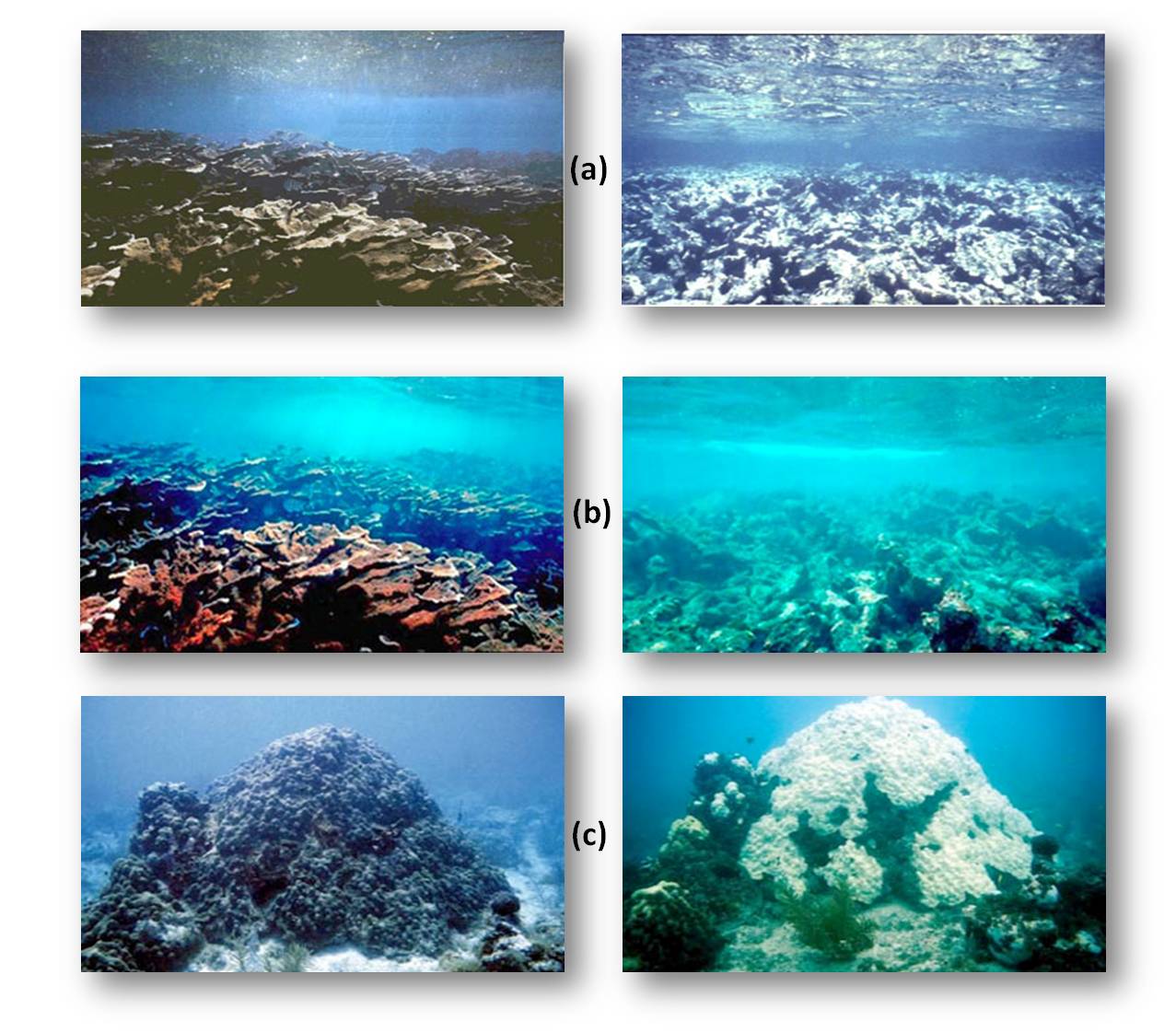}
		\captionsetup{justification=centering,margin=2cm}
	\caption{Examples of coral reef decay: (a) Coral ‘Elkhorn’ in the Caribbean sea (1975-1995) \cite{YipOnline}, (b) ‘Carysfort’ reef in the Florida Keys (1975-2004) \cite{DunstanOnline}, (c) Bleaching of 500 years old coral head (1996-1997) \cite{Quirolo2010}}
	\label{fig:Coral_Decay}
\end{figure}

Based on mid-90's statistics \cite{ICRANOnline}, 10\% of coral reefs were destroyed and can't be recovered again, and there are only less than 30\% healthy coral reefs around the world. Figure~\ref{fig:0302-coralreefs-EN} shows human activities that threaten the coral reefs around the world starting from Caribbean coast of Atlantic ocean as presented in figure~\ref{fig:Coral_Decay}(a,b); passing through east Africa coast \& Red sea; ending to central part of Pacific ocean, those activities involve coastal development (sun blocking from eroding soil over aquatic world), underwater pollution (oil, gas, and mineral exploration \& extraction), destructive fishing methods (very popular in south Pacific and southeast Asia using poison fishing and dynamite fishing), and unsustainable tourism (i.e. touching during diving sessions), such that they damage both cold deep and warm shallow coral physically and don't allow them to grow again or recover in decades \cite{Diop2002,De2012}.

There two more major coral environmental-based threads \cite{KleypasOnline}: (coral bleaching, and ocean acidification). 16\% of the world's coral reefs is suffered from first thread over the last three decades  due to increase in water temperature which causes losing coral color and become white as shown in figure~\ref{fig:Coral_Decay}(c), but corals can recover from bleaching. Unstable balance in atmospheric materials (i.e. increased carbon dioxide) leads to later threat which causes lowering of ocean pH (acidity measure) and affect coral negatively by losing their calcium selections.
There two more major coral environmental-based threads: coral bleaching, and ocean acidification. 16\% of the world's coral reefs is suffered from first thread over the last three decades  due to increase in water temperature which causes losing coral color and become white as shown in figure~\ref{fig:Coral_Decay}(c), but corals can recover from bleaching. Unstable balance in atmospheric materials (i.e. increased carbon dioxide) leads to later threat which causes lowering of ocean pH (acidity measure) and affects coral negatively by losing their calcium selections.

\section{Coral Transplantation}

Some types of coral reef have a slow survival ability for recovering or re-growth using small healthy coral wreckage resulting their artificial coral colony after some decades. Possible strategies are provided for coral gardening through involvement of SCUBA divers in coral reef reassemble and transplantation. Although, some limitations (restricted time and depth per diving session respect to human abilities) are introduced a small survival rate in transplanted corals (especially cold sea corals due to their deep depth conditions). Coral ecologists  investigate new robot-based strategy in deep-sea coral restoration in such that autonomous underwater vehicles (AUVs) grasp cold-water coral samples and replant them in damaged reef areas. Successful transplantation trail \cite{Roberts2009} is already occurred in 2008 for cold-water coral Lophelia (at 82m water depth) in Kosterfjord, Sweden. 

\begin{figure}
	\centering
		\includegraphics[width=1.00\textwidth]{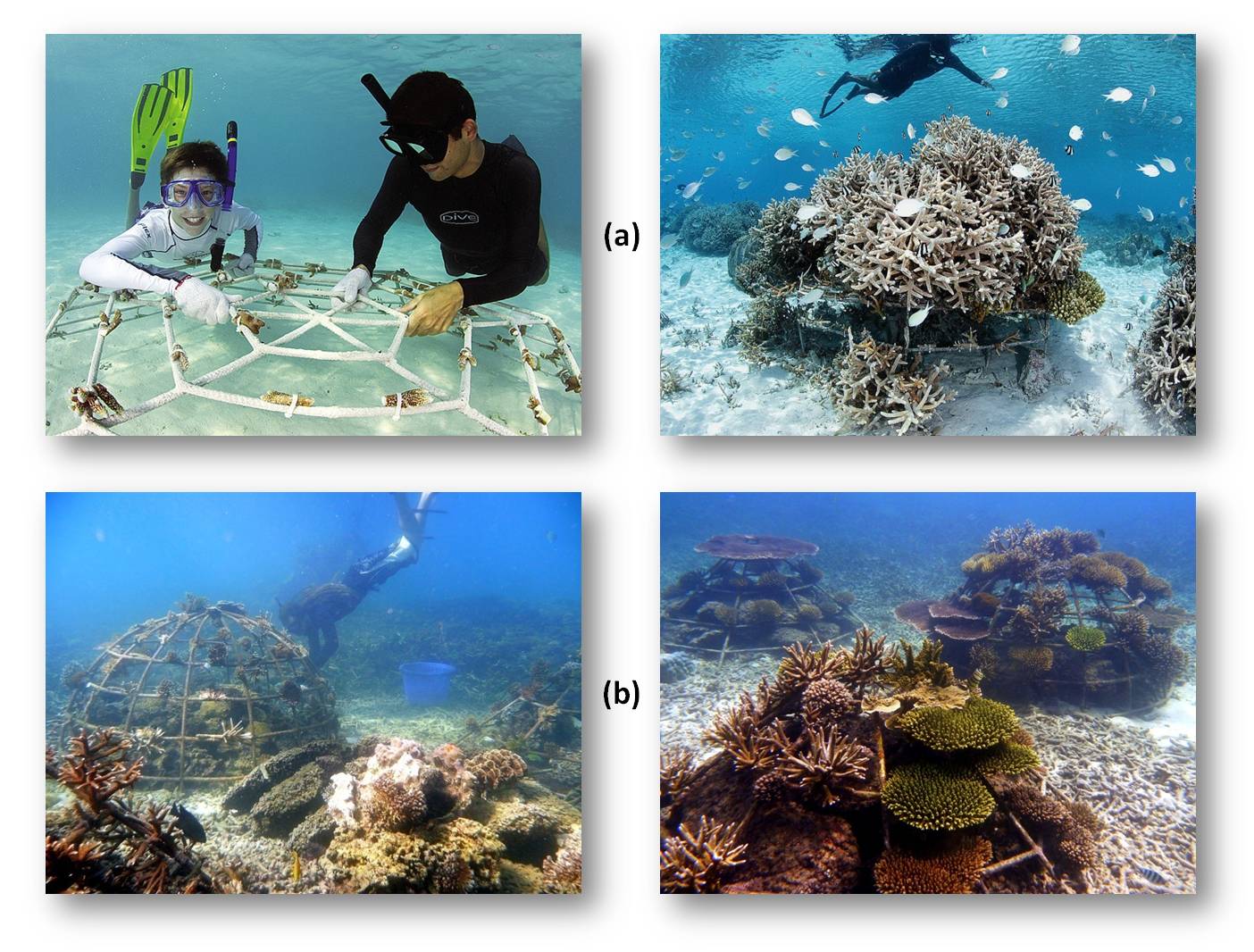}
		\captionsetup{justification=centering,margin=2cm}
	\caption{Examples of coral reef transplantation: (a) Reviving coral reefs in the Maldives (Reefscapers Project 2001) \cite{ReefscapersOnline}, (b) Rehabilitation of coral reefs in Koh Tao island, Thailand (Save Coral Reefs 2012) \cite{KohTaoOnline}}
	\label{fig:Coral_Trans}
\end{figure}

Figure~\ref{fig:Coral_Trans} shows two examples of human-based transplantation for coral reef fragments. problem of first example start in 1998 when increase in ocean temperature due to storms cause coral bleaching and lose 90\% of shallow corals. World-known resort (Four Seasons) and environmental consultancy agency (Seamarc) started a coral-saving project entitled (Reefscapers) in 2001 to transplant coral fragments to artificial coral reefs and monitor their growth over years; leading to amazing results (20\% coral increase, 80\% survival rate for transplanted corals). global warming caused second example 80\%	coral death at Koh Tao (Thailand) in 2010, so that an environmental organization (Save Coral Reefs) begin a coral restoration project resulting a tremendous coral growth after one year of operation.

\section{Autonomous Underwater Vehicles}

Although deployment of single AUV operation is time limited. Inspired from behavioral of natural swarms of insects (bees, wasps and termites) in building complex colonies, team of marine biologists and robotics experts introduced an innovative underwater project 'coralbots' to speed-up the regenerated coral process using intelligent swarms of inter-connected AUVs.

\begin{figure}
	\centering
		\includegraphics[width=1.00\textwidth]{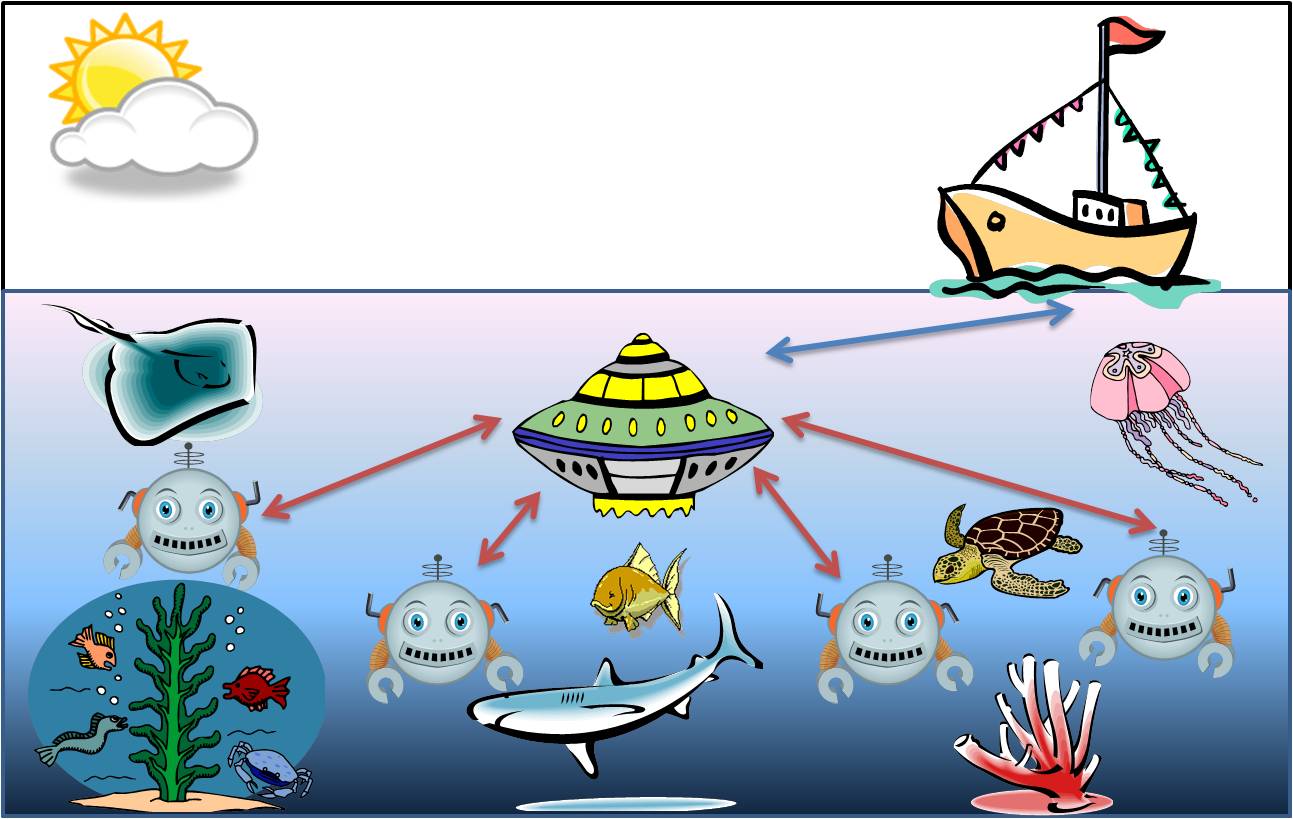}
	\caption{Retransplantation process of autonomous underwater robots (AUVs) for coral reefs}
	\label{fig:UAVs_Coral}
\end{figure}

Proposed work (as shown in figure~\ref{fig:UAVs_Coral}) consists of two stages: offline data training and online identification. offline training will be on surface workstation for fast computation and long execution time which extracts features from coral-labeled images besides spatial information and then apply deep learning process (supervised, unsupervised, or hybrid) to get well-trained parameters for further successful classification. However, online identification will be on remotely operated underwater vehicle (ROV) which collect images from several AUVs and find out which species are included, and detect their coordinates in real-time processing for further coral transplantation.

\chapter{State of the art} \label{chap:state}

This chapter discusses most-recent research in classification for coral species using optical camera sensors, then explains convolutional neural networks (deep learning method) as a feature extraction and classification technique and its successful applications in related research (object classification, and object recognition \& detection).

\section{Coral Classification} \label{sect:coral_class}

Clement et al. \cite{Clement2005} presented local binary pattern (LBP) as feature descriptor for binary detection of crown-of-thorns starfish (COTS) from great barrier reef images in Australia, in which his experiments achieved an above-average results in only one-class segmentation. Mehta et al. \cite{Mehta2007} used support vector machines (SVM) classifier directly on spatial information of coral image without any preprocessing step to get a binary (coral/non-coral) output, he reasoned for using raw-data as classifier input such that features and descriptors for coral textures are very difficult to obtain. he achieved 95 \%  correct classification but his method behaves negatively with any change in underwater illumination.

Pizarro et al. \cite{Pizarro2008} introduced object recognition for coarse habitat classification (1st experiment; 8 classes: coralline rubble, hard coral, hard coral +soft coral + coralline rubble, halimeda + hard Coral + coralline rubble, macroalgae, rhodolith, sponges, and un-colonized) (2nd experiment; 4 classes: reef + coarse sand, coarse sand, reef, and fine sand), he employed the same color features of Marcos but different texture features based on bag-of-words using scale-invariant feature transform (SIFT) with extra saliency feature of gabor-filter response, this method can be only used with single-object images (one classes per entire image) \cite{Shihavuddin2013a}. Marcos et al. \cite{Marcos2008} developed an automated rapid classification (5 classes: coral, sand, rubble, dead coral, and dead coral with algae) for underwater reef video, he used color features based on histogram of normalized chromaticity coordinates (NCC) and texture features from local binary patterns (LBP) descriptor, those features feed into linear discriminant analysis (LDA) classifier. in case of using more classes \cite{Shihavuddin2013a}, his method output inaccurate classification.

Johnson-Roberson et al. \cite{Johnson-roberson2006} showed an approach for the autonomous segmentation and classification of coral through the combination of visual and acoustic data, which generates 60 visual features: 12 are the mean and standard deviation of all of the RGB and HSV channels separately and the remaining 48 are obtained by convolving the region with Gabor wavelets at six scales and four orientations and taking the mean and standard deviation of the results for each scale and orientation combination, then SVM is selected for classification task, pre-processing is mainly required to allocate foreground regions before feature extraction. Purser et al. \cite{Purser2009} investigated machine-learning algorithms for the automated detection of cold-water coral habitats, which computes 15 differently oriented and spaced gratings in order to produce a set of 30 texture features, and to compare a computer vision system with the use of three manual methods: 15-point quadrat, 100-point quadrat and frame mapping. Strokes \& Deane \cite{Stokes2009} described an automated algorithm for the classification of coral reef benthic organisms and substrates which divides image into blocks, then finds distance between those blocks and identifies species blocks based on color features (normalized histogram of RGB color space) and texture features (radial samples of 2D discrete cosine transform) by using inconvenient distance metric (manually assigned parameters) after unsuccessful results of well-known mahalanobis distance.

Beijbom et al. \cite{Beijbom2012} introduced Moorea Labeled Corals (MLC) dataset and proposed multi-scale classification algorithm for automatic annotation, he developed color stretching for each channel individually in L*a*b* color space as pre-processing step, then used Maximum Response (MR) filter bank approach (rotation invariant) as color and texture feature, followed by applying Radial Basis Function kernel (RBF) of Support Vector Machines (SVM) classifier, this method seek all possibilities (time-consuming) to find a suitable patch size around selected image points for species identification. 

Schoening et al. \cite{Schoening2012} introduced semi-automated detection system for deep-sea coral images, in which firstly applies preprocessing step for illumination correction, then secondly extracts high dimensional features at labeled pixels based on MPEG7 standard (four descriptors for color features; three for texture; ten for structure and motion), after that a set of successive different support vector machines is applied a long side with thresholding post processing, although the used features are more generic for any application leading to high sensitivity for underwater cluttering background. Stough \cite{Stough2012} presented an automatic binary segmentation technique of live Staghorn coral species, in which regional intensity quantile functions (QF) is used as color features and Scale-Invariant Feature Transform (SIFT) is maintained to be texture features, followed by linear SVM as classifier, this supervised technique is highly noise sensitive (due to SIFT).

Shihavuddin et al. \cite{Shihavuddin2013a} implemented hybrid variable-scheme classification framework for benthic coral reef images or mosaics, this framework uses combination of the following features (local binary pattern (CLBP), grey level co-occurrence matrix (GLCM), Gabor filter response, and opponent angle and hue channel color histograms) and also combination of the following classifiers (k- nearest neighbor (KNN), neural network (NN), support vector machine (SVM) or probability density weighted mean distance (PDWMD)) along besides middleware procedures for better result enhancement, but this framework works sufficiently with small-patched images (high negative impact of background information).

\begin{table}
\begin{center}
\includegraphics[width=\textwidth,height=\textheight,keepaspectratio]{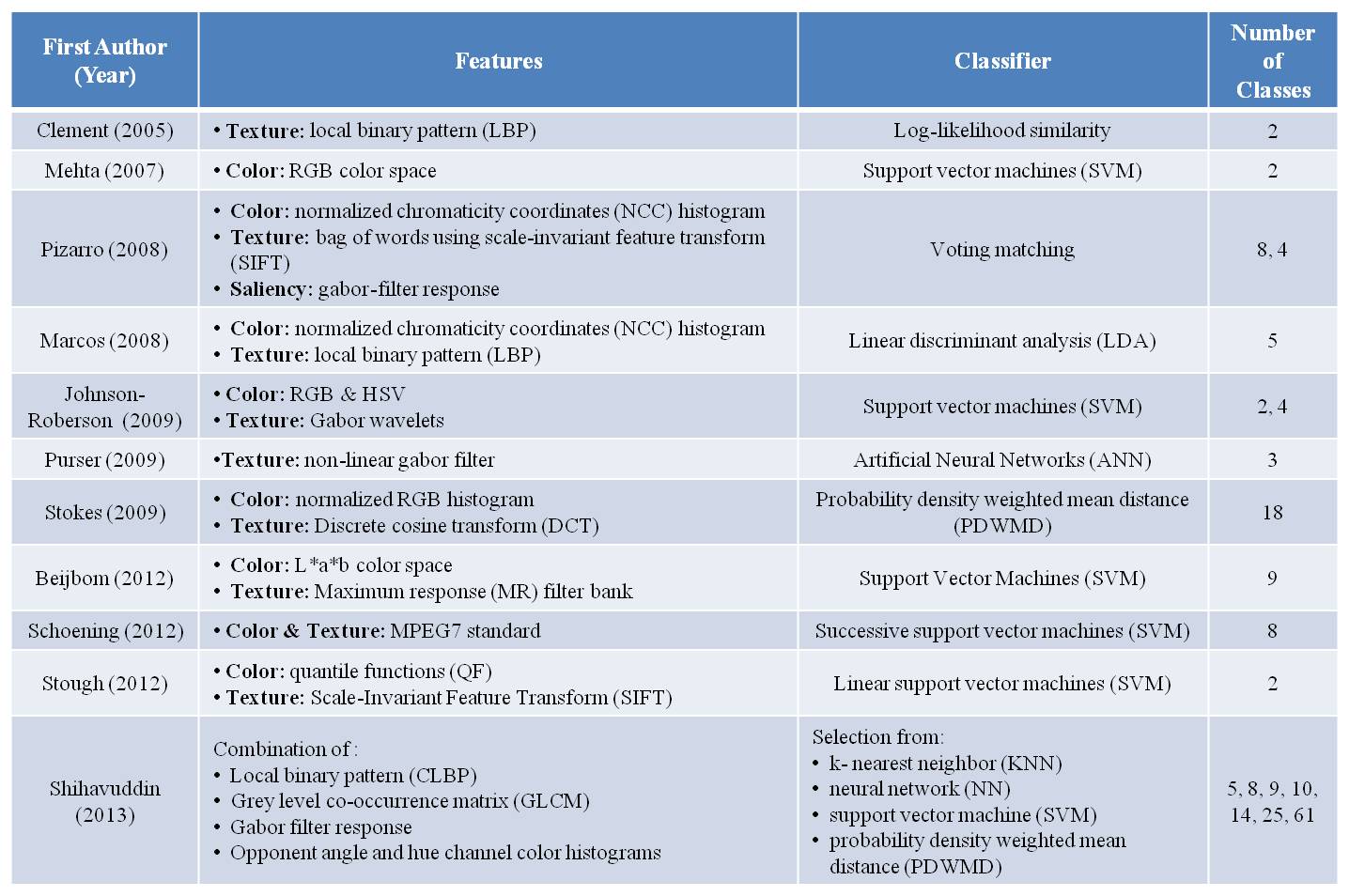}
\caption{Summary of related methods for coral image classification}
\label{table:related_coral} 
\end{center}
\end{table}

Rather than depending on human-crafted features (please see table~\ref{table:related_coral}) to get a proper coral classification, the proposed work decides letting the feature mapping to be done automatically by deep convolutional neural networks regardless to any under-water environment condition. by feeding new images, the network can learn and adapt the constructed feature maps respect to desired class outputs.

\section{Convolutional Neural Networks} \label{sect:CNN}

\begin{figure}
	\centering
		\includegraphics[width=1.00\textwidth]{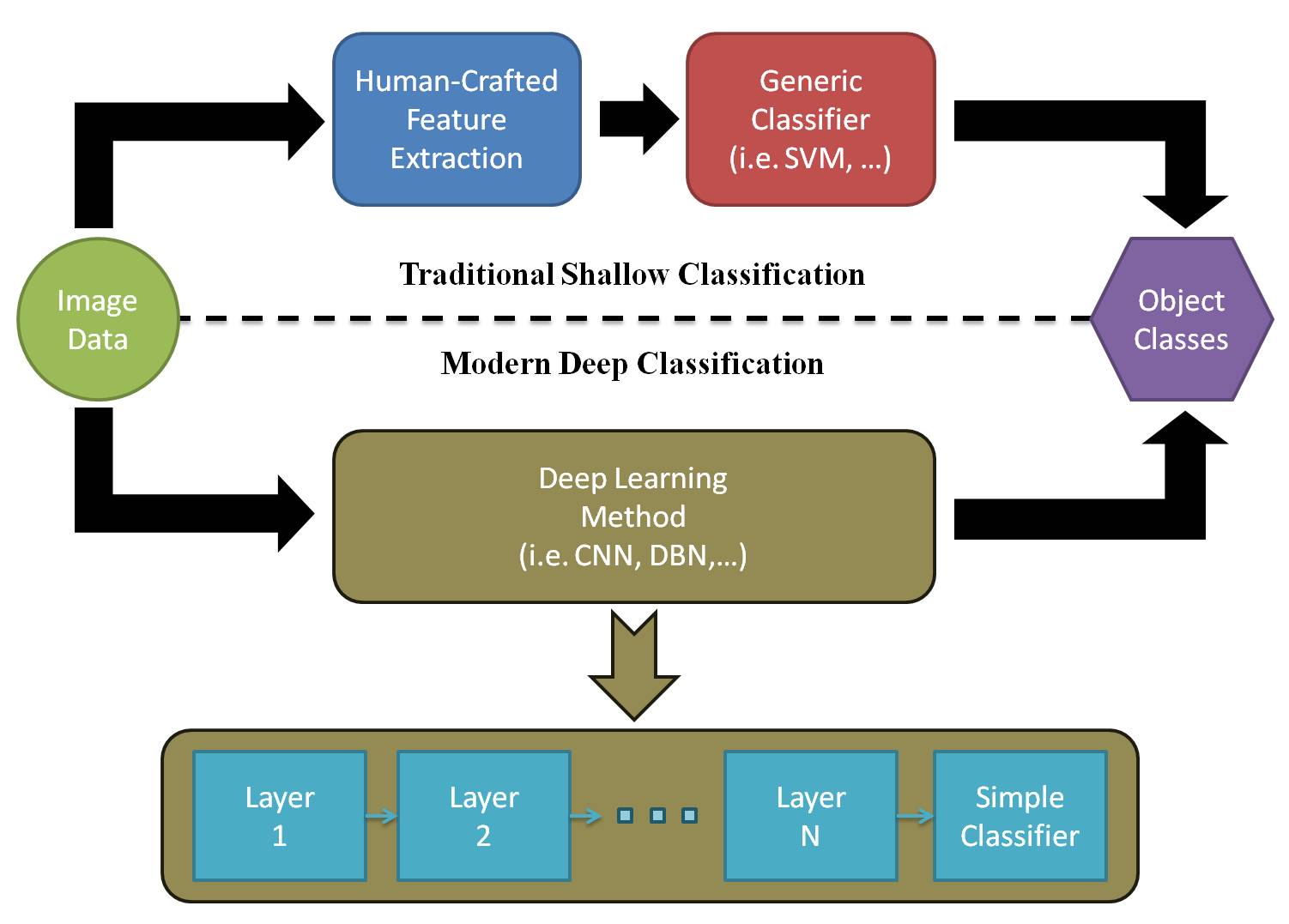}
		\captionsetup{justification=centering,margin=2cm}
	\caption{Difference between shallow “traditional” and deep “modern” classification architectures}
	\label{fig:ShallowVsDeep}
\end{figure}

Traditional architecture firstly extracts hand-designed key features based on human analysis for input data, secondly applies those features in form of data vectors to generic classifier in order to get predicted target classes (in other words, classifier is totally dependent on how features constructed not input data). Deep architecture trains learning features across hidden layers; starting from low level details (i.e. edges, corners) up to high level details (i.e. shape, texture); to get better data representation for simple classifier (please see figure~\ref{fig:ShallowVsDeep} for graphical details).

\begin{figure}
	\centering
		\includegraphics[width=1.00\textwidth]{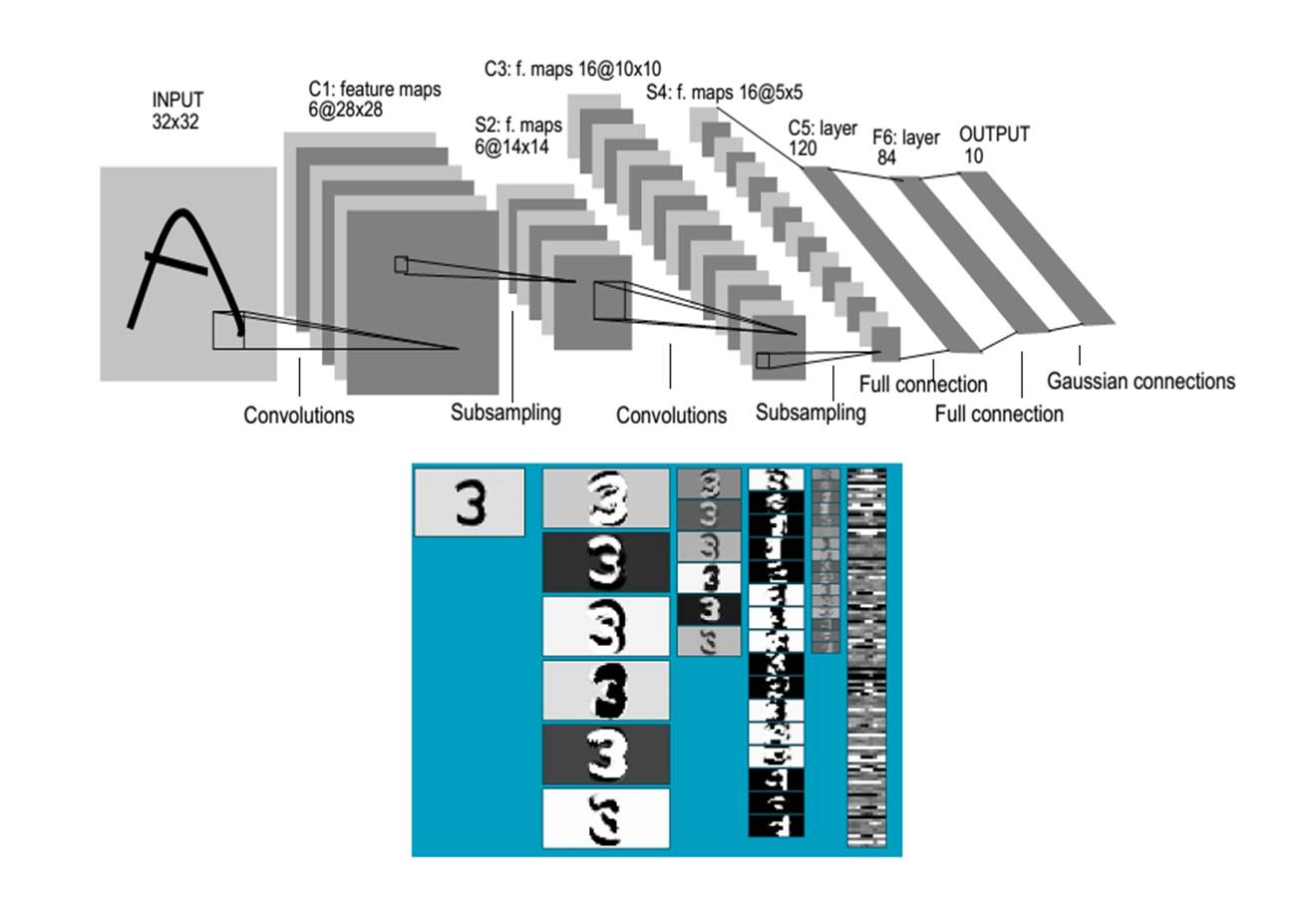}
		\captionsetup{justification=centering,margin=2cm}
	\caption{Architecture of LeNet-5 (Convolutional Neural Networks) for digit recognition, from LeCun\cite{LeCun98}}
	\label{fig:CNN_LeNet5}
\end{figure}

A convolutional neural network (CNN) \cite{Hubel68,MatsuguMMK03,LeCun2012} is a type of feed-forward back-propagation neural networks respect to biological-based visual processes. it consists of trainable multiple convolutional stages \cite{Palm2012}, in which input and output of each stage are variant representation of one/multi-dimensional array (i.e. 1D for audio, 2D for image, 3D for video, ...), the output array learn to extract high-receptive features from all sides of input one. A typical CNN is composed of two or three stages, followed by a classification layer. LeCun presented first back-propagation CNN entitled "LeNet-5" (please see figure~\ref{fig:CNN_LeNet5}) for handwritten digit recognition, which is a large network which contains 6 layer hidden layers whose its input is 28x28 input image of single hand-written character and its output is multi-invariant feature map of input character.

\begin{figure}
	\centering
		\includegraphics[width=1.00\textwidth]{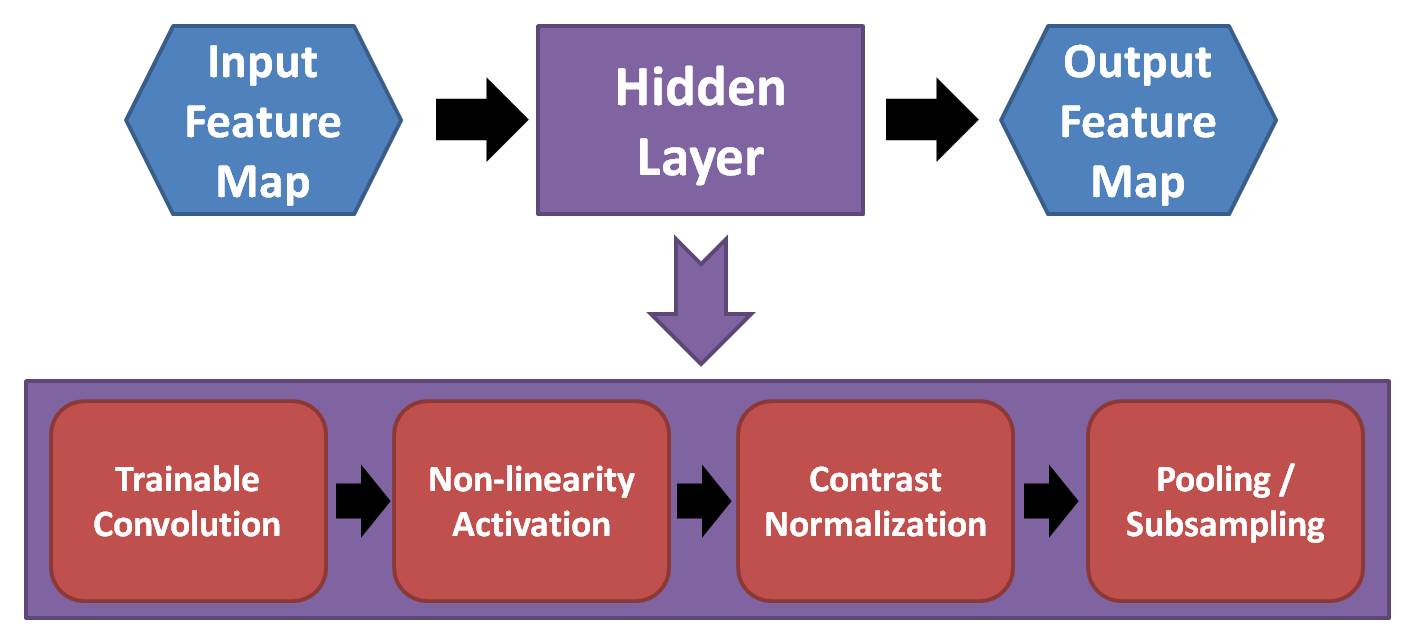}
	\caption{Main structure of CNN hidden layer}
	\label{fig:CNN_Hidden}
\end{figure}

Each hidden stage/layer consists of four steps (as shown in figure~\ref{fig:CNN_Hidden}) : trainable convolution, non-linearity activation, contrast normalization, and pooling/sub-sampling. Convolution filters an input map into translation-invariant maps with different trainable weights and biases, Non-linear activation function (i.e. hyperbolic, sigmoid, \ldots) adds independent relationship within objects inside, contrast normalization keeps output maps in pre-defined range measures, final feature map is subsampled or max-pooled from output maps of the last stage (make it small in size to further faster calculation in next layers).

\subsection{Object Classification}
Buyssens \cite{Buyssens2012} introduced multi-scale convolutional neural networks (MCNN) for cancer cell image classification, in which cells are detected and segmented from virtual slides, then they are classified by using different CNNs at each scale and finally fuse their outputs through linear combination to get six different types of cancer cells. activation function used at convolution layer  is non-linear squashing function, output of max-pooling layer is calculated in selecting regions of maximum activation instead of averaging regions (classical way). This method achieved lower error rate (5.74\%) than state-of-art approaches after 50 epochs with four different square scales (80, 56, 40, 28).\\

Krizhevsky \cite{Krizhevsky2012} presented a large deep convolutional neural network with eight hidden layers to achieve state-of-art results (1.5 million images of 256x256 down-sampled size) classification contest (LSVRC2010, LSVRC2012 of 37.5\% and 36.7\% error rates respectively with one thousand different classes), in which second best result from other participants has at least 10\% error rate than  proposed work. Such kind of large network faced over fitting over huge number of epochs which can be solvable by (1) introducing new transformed data from original data (2) dropping out neutral hidden neurons.

\subsection{Object Recognition and Detection}

\begin{figure}
	\centering
		\includegraphics[width=1.00\textwidth]{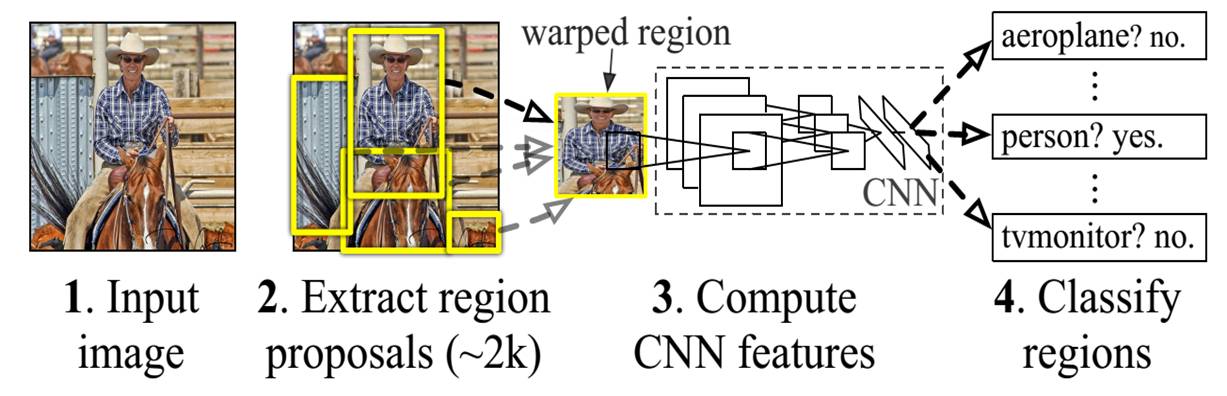}
	\caption{Overview for object detection using Regions with CNN features (R-CNN\cite{Girshick2013}) }
	\label{fig:RCNN}
\end{figure}

Girshick \cite{Girshick2013} constructed innovated object detection (as shown in figure~\ref{fig:RCNN}) with state-of-art performance (mean average precision = 53.3\%) on PASCAL Visual Object Classes Challenge 2012 over complex methods of different combination of low level features. He used selective search for extraction of candidate regions and proposed region-based convolutional neural networks to classify the target objects using supervised training and detect them using domain-specific fine-tuning process. Syafeeza \cite{Syafeeza2014} proposed a face recognition system using convolutional neural network with over 85\% accuracy in two different datasets. non-linear data representations (changes in illumination and poses) are overcome by experimenting variant network adjustment (fusion architecture, 10-fold cross validation, partial inter-layer connection, hyperbolic tangent activation function, input data normalization, and Gaussian weight normalization).

\begin{figure}
	\centering
		\includegraphics[width=1.00\textwidth]{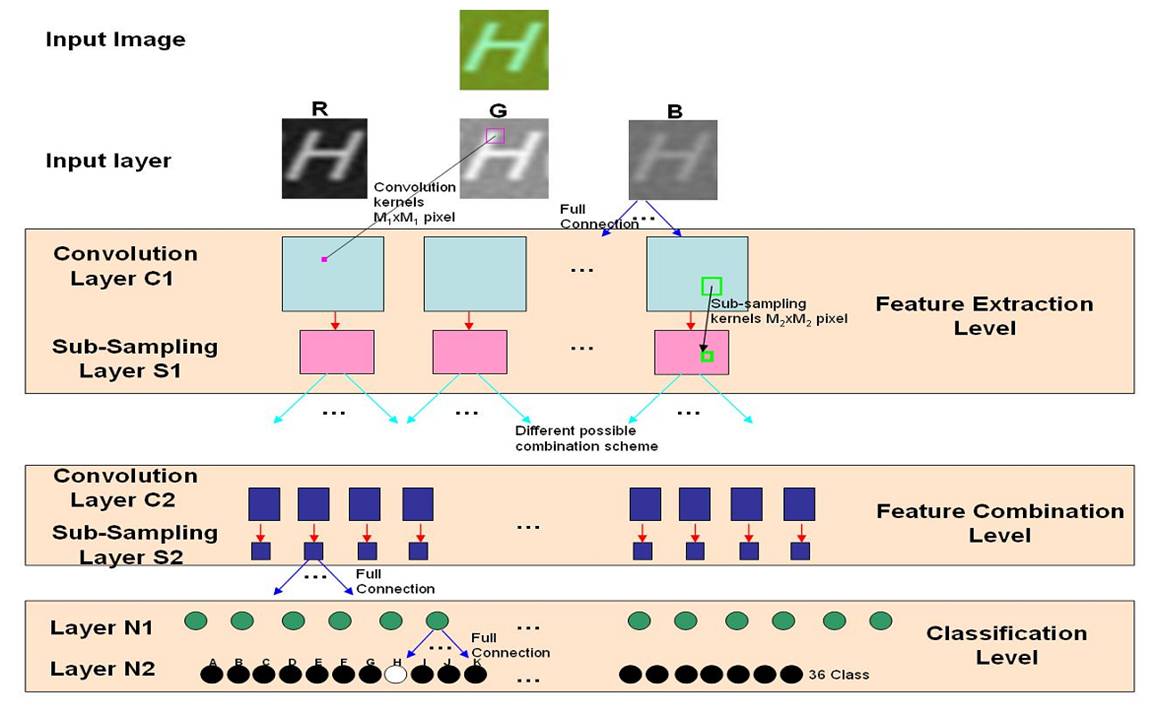}
	\caption{Network architecture for text recognition \cite{Saidane2007}}
	\label{fig:CNN_TextRecog}
\end{figure}

Pinheiro \cite{pinheiro2014} used recurrent-based convolutional neural networks to successfully label each pixel of input images in the following datasets: Stanford background (80.2\% accuracy, 715 images, 8 classes, 320x240 pixels) and SIFT flow dataset (77.7\% accuracy, 2688 images, 3 classes, 256x256 pixels). These results exceed state-of-art ones in both datasets with efficient computation time. Saidane \cite{Saidane2007} showed a robust recognition algorithm (as shown in figure~\ref{fig:CNN_TextRecog}) for color text characters against different types of noises (high-detailed background, non-uniform lighting, etc\ldots). recognition rate 84.5\% is achieved with 36 classes (26 alphabetic letters and 10 numerical digits) using ICVDAR 2003 database.

\chapter{Methodology} \label{chap:method}
This chapter presents mains steps of coral classification based on convolutional neural networks, and overview \& implementation details of proposed classification framework to match coral image points to target coral classes.

\section{Introduction}

Coral-reef classification can be divided into three main consecutive steps:
\begin{enumerate}
\item Under-water image de-noising (preprocessing step): due to different challenges (motion blurring, color attenuation, refracted sunlight patterns, water temperature variation, sky color variation, scattering effects, presence of sea particles, etc...),  raw image must be enhanced visually to show its coral species in details for further steps.
\item Feature extraction: image which contains different coral species, have to find salient regions in each object in order to identify and distinguish those species easily respect to invariance of the following aspects (illumination, rotation, size, view angle, camera distance, etc...).
\item Machine Learning (ML) algorithm: those extracted features are used as input for machine learning to find suitable parameters to converge species of new images to similar trained ones respectively.
\end{enumerate}
The proposed framework implemented the first step of coral reef classification (under-water image de-noising) using state-of-art research in underwater color enhancement and some additional data adjustment (hybrid sized patching) due to point-based data representation, and also achieved the second and third steps together using convolutional neural networks (deep learning method) along side with construction of salient feature maps for faster network convergence based on most recent techniques in computer vision.

\section{Framework Overview}

\begin{figure}
	\centering
		\includegraphics[width=1.00\textwidth]{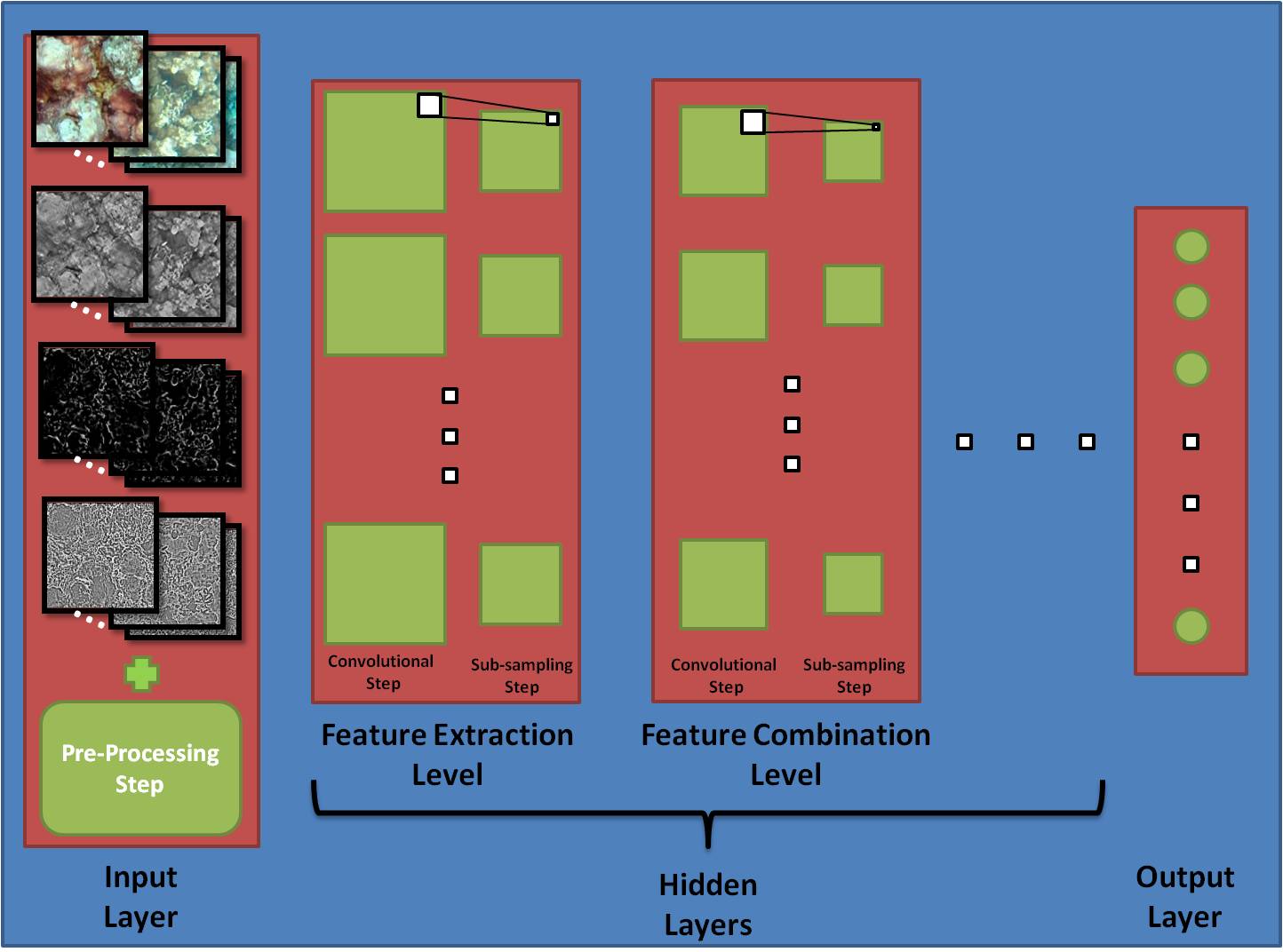}
	\caption{Architecture overview of proposed CNN}
	\label{fig:ProposedCNN}
\end{figure}

\begin{figure}
	\centering
		\includegraphics[width=1.00\textwidth]{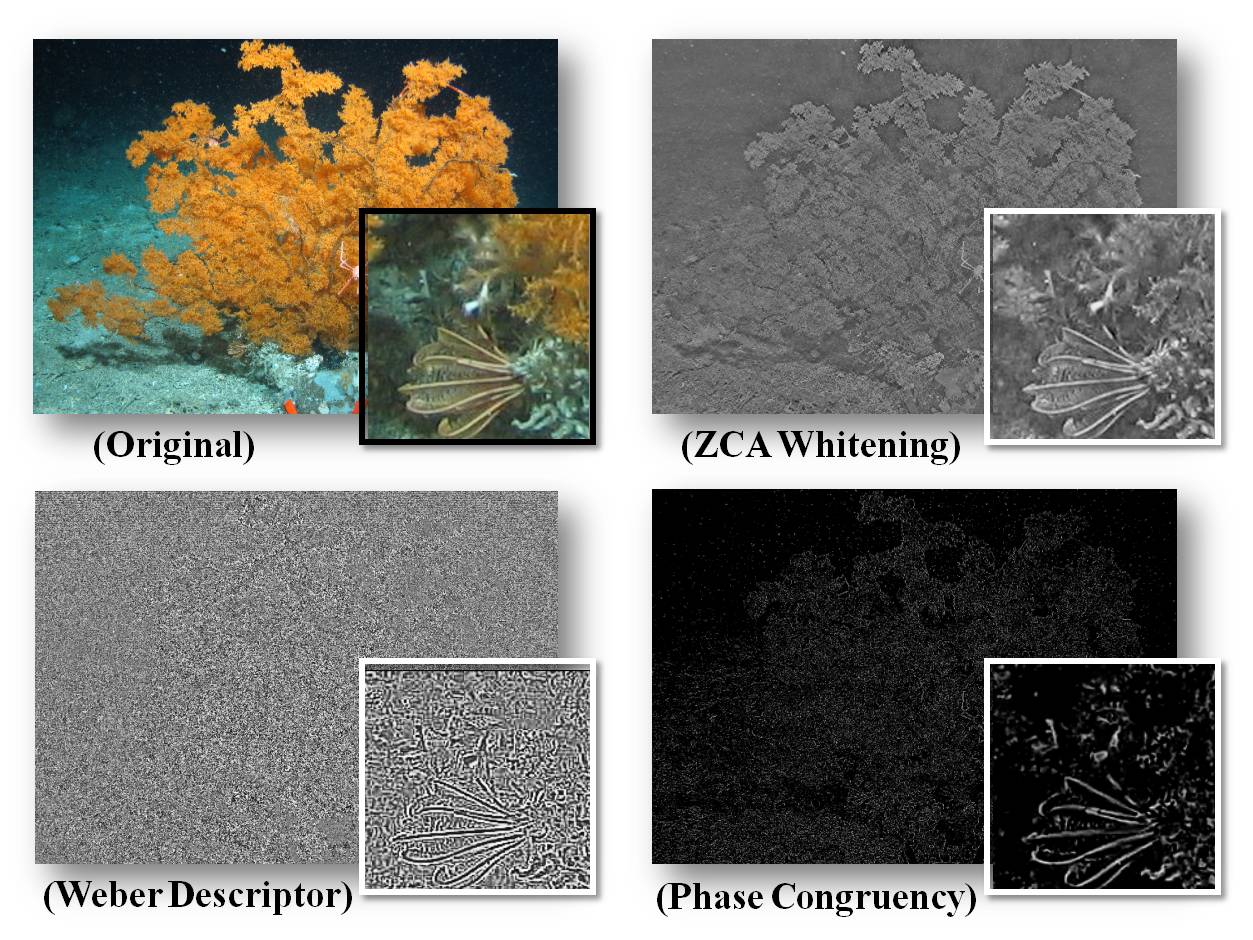}
		\captionsetup{justification=centering,margin=2cm}
	\caption{Example of feature maps for Crinoid coral using ADS dataset}
	\label{fig:Feature_Deep}
\end{figure}

\begin{figure}
	\centering
		\includegraphics[width=1.00\textwidth]{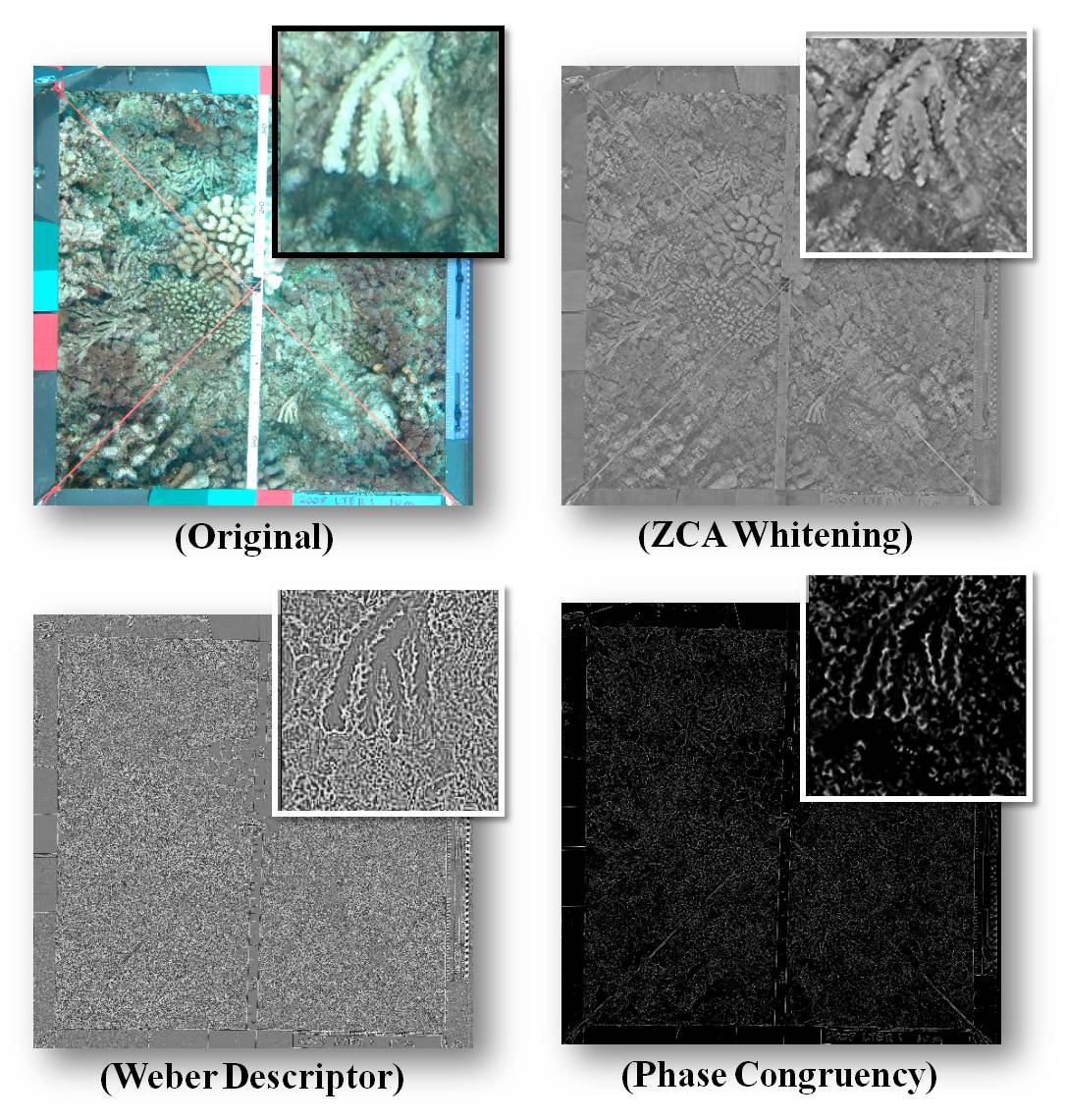}
		\captionsetup{justification=centering,margin=2cm}
	\caption{Example of feature maps for Acropora coral using MLC dataset}
	\label{fig:Feature_MLC}
\end{figure}

The proposed classification framework (as shown in figure~\ref{fig:ProposedCNN}) contains three main levels (input layer, hidden layers, output layer). Input layer consists of three basic channels of color image plus extra channels for texture and shape descriptors consisting of following components: zero component analysis whitening, phase congruency, and Weber local descriptor (as shown in figures~\ref{fig:Feature_Deep},~\ref{fig:Feature_MLC}), and preprocessing step (color correction/enhancement, smoothing filter) can be applied for further classification improvement. Hidden layers contains one or more layer(s) [usually 2 or 3] in which each layer consists of convolution layer followed by down-sampling layer in such way that the network can find suitable weights of convolutional kernel and additive biases. Almost first layer represents feature extraction by finding visual strokes, edges, and corners, and up-coming layers starting from second layer show how those features can combine in different aspects to get a discriminative output map for each target class. Output layer acts as a classification layer and symbolize reconstructed maps from last hidden layer into binary vector (placement of number one in specific element corresponding to desired class and number zero in the rest).

\section{Implementation}

\subsection{Preprocessing}
\subsubsection{Color Enhancement}

Bazeille \cite{Bazeille2006} discussed difficulties in capturing good quality under-water images due to non-uniform lighting and underwater perturbation, he introduced a large parameter-free algorithm consisting of applying set of different filters into noisy underwater images which enhances edges and visual quality robustly. Iqbal \cite{Iqbal2007} addressed under-water lighting problems due to light absorption, vertical polarization, and sea structure in which short wavelength of blue leads it to penetrate into sea layers and be a dominant color in deep water, he presents a simple slide color stretching algorithm based on RGB and HSI color models that is efficient equalization for color contrast. Beijbom \cite{Beijbom2012} stated compensation of color differences in underwater turbidity and illumination can be solved by simply stretching histogram of each color channel separately with respect to 1\% and 99\% intensities.

\begin{figure}
	\centering
		\includegraphics[width=1.00\textwidth]{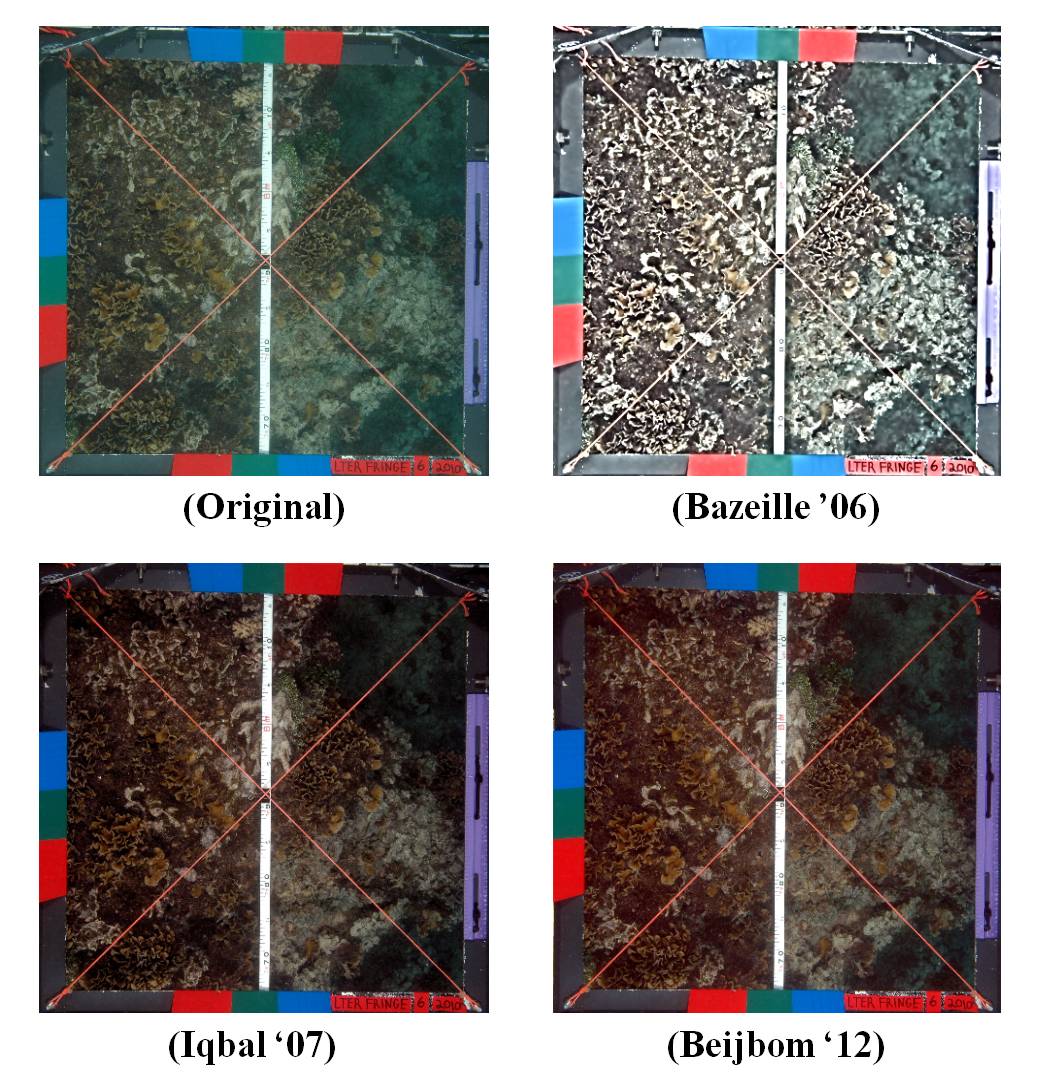}
	\caption{Example of color enhancement for coral images}
	\label{fig:ColorEnhance}
\end{figure}

Figure~\ref{fig:ColorEnhance} presents state of art coral processing for color enhancement and its application on un-cleared coral image from MLC dataset, in which bazeille work shows edge-based version with less-coloring objects, Iqbal presents well-cleared version for foreground objects, and beijbom presents same output of Iqbal but with more reddish color on corals (as shown in left side of image).

\subsubsection{Hybrid Patching}

\begin{figure}
	\centering
		\includegraphics[width=1.00\textwidth]{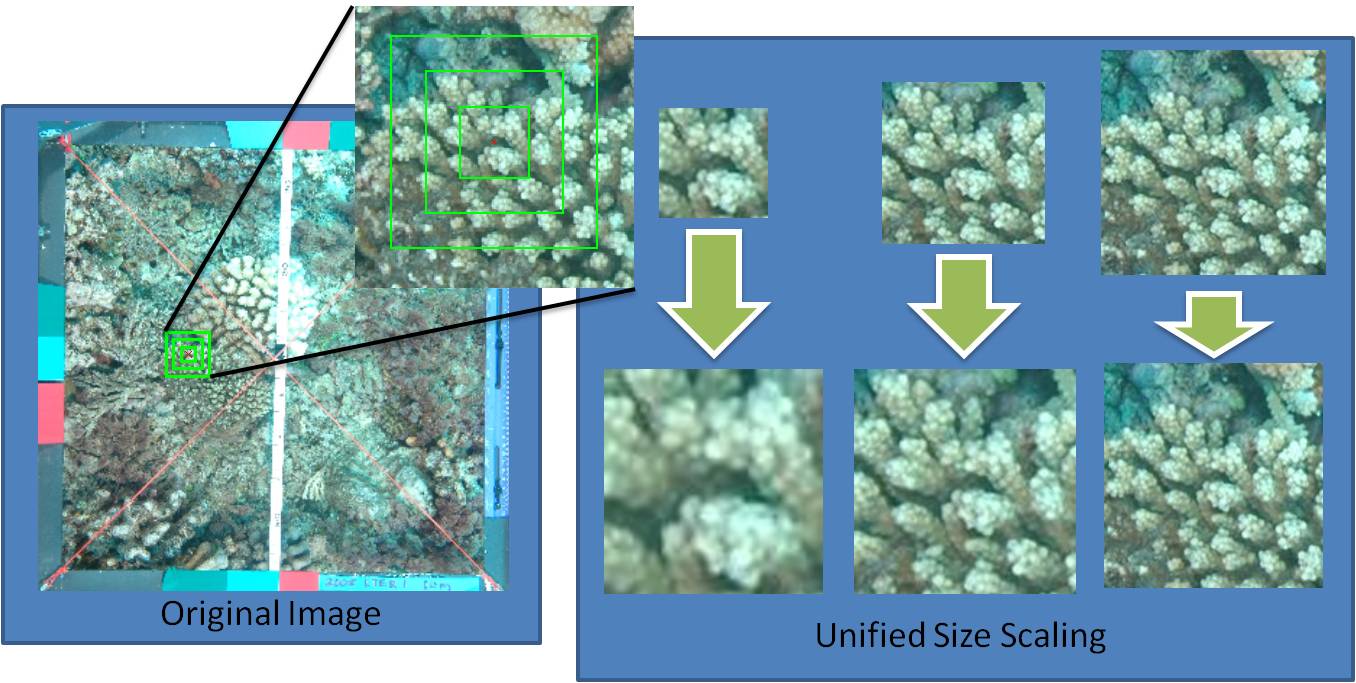}
	\caption{Example of hybrid patching}
	\label{fig:HybridPatching}
\end{figure}

Three different-in-size patches are selected across each annotated point (61x61, 121x121, 181x181), then unified size scaling step is applied to those patches by scaling them up to size of the largest patch (181x181) allowing pixel randomization (blurring) in inter-shape coral details and keeping up coral’s edges and corners (please see figure~\ref{fig:HybridPatching}), or scaling them down to size of the smallest patch (61x61) for fast classification computation over small data representation of different scaling selections.

\subsection{Feature Maps}
\subsubsection{Zero Component Analysis Whitening}
Zero Component Analysis (ZCA) whitening \cite{Bell97,MemisevicOnline,NgOnline,DelormeOnline} makes data less-redundant by removing any neighboring correlations in adjacent pixels in such that output data removes amplitude information and keeps recognizable edges. it stimulates image scanning retinal process which decorrelates similar intensity values of contiguous pixels (high correlated adjacent pixels) after few moments of eye-focusing. It requires one smoothing parameter (very small number) preventing division of zero in its calculation with respect to tiny eigenvalues which leads to a better-visual output features (dispatching off the inter-process aliasing artificats).

\begin{figure}
	\centering
		\includegraphics[width=1.00\textwidth]{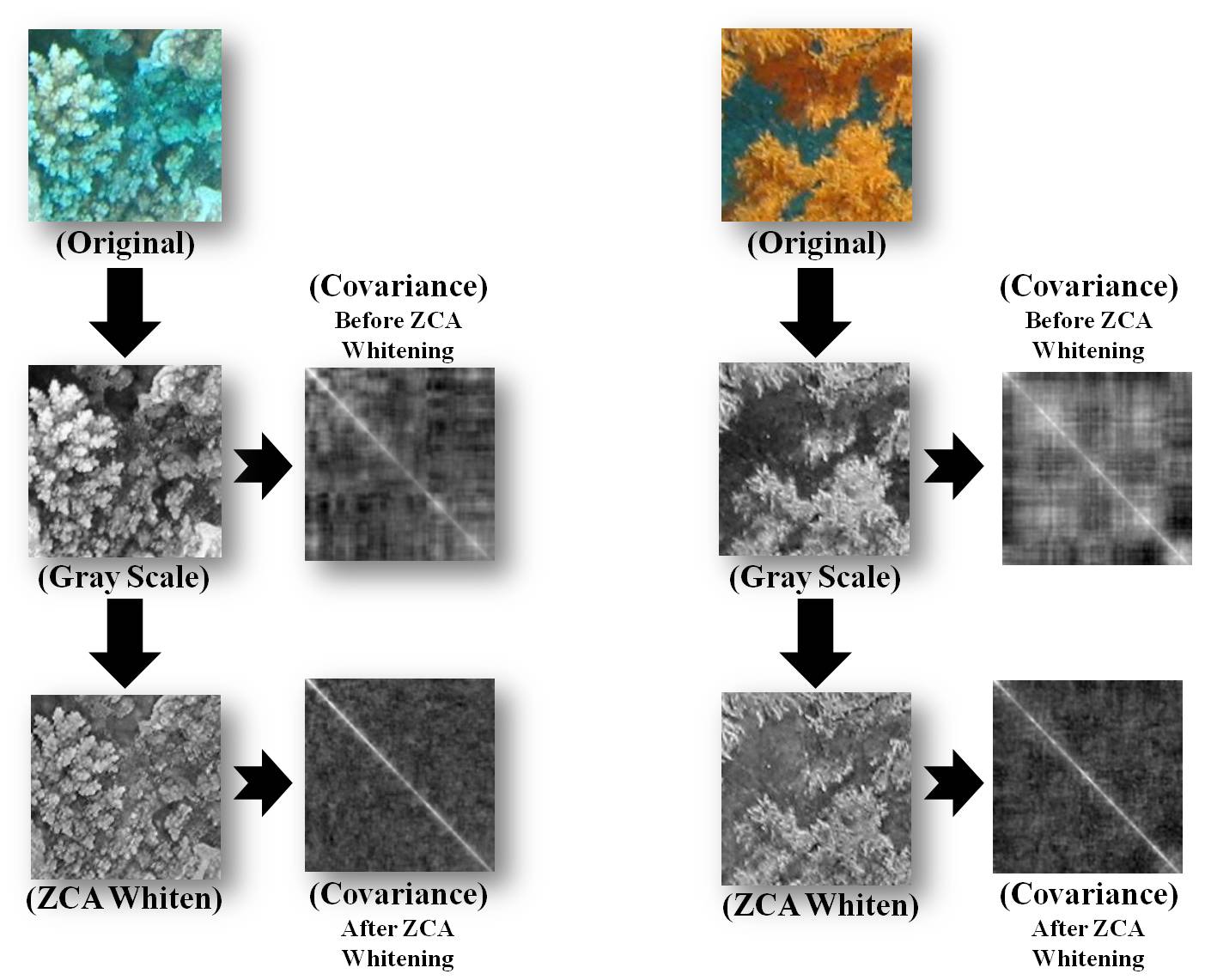}
	\caption{Two examples of ZCA whitening using MLC and ADS dataset}
	\label{fig:ZCA_Whitening}
\end{figure}

Figure~\ref{fig:ZCA_Whitening} shows how image data are correlated before and after ZCA whitening (covariance matrix represents correlation between image rows, in which diagonal white value represents full correlations of rows with itself and semi-correlated with rows with other rows). Before ZCA whitening, high correlation is found in covariance of grayscale version between adjacent rows (white blobs across main diagonal) and less correlation between faraway rows (black blobs across cross diagonal). After ZCA whitening, output image is polished by discarding inter-coral details and focusing on external shape of coral, wherefore homogenous correlation between non-duplicating rows exists in its covariance (everywhere except main diagonal).

\subsubsection{Weber Local Descriptor}

Weber Local Descriptor (WLD) \cite{Chen2010} is inspired from psychological law in 19th century ``Weber's Law" and represents human perception of a pattern depending on ratio between change in image pixel and original pixel value. it consists of two components: differential excitation and orientation. The differential excitation component computes the salient micro-patterns relative to nearby neighboring pixels by calculating a function of the ratio between ratio between the relative intensity differences of a current pixel against its neighbors and the intensity of the current pixel. The orientation component constructs statistics on the computed salient patterns along with the gradient orientation of current pixel by building histograms of dominant orientations. This method shows a robust edge representation of high-texture images against high-noisy changes in illumination of image environment. WLD has proven promising results in different object recognition issues \cite{PalDSGN13,GhulamHAMBA12,LiGY13}.

\subsubsection{Phase Congruency}
Phase Congruency (PC) \cite{Morrone1988,Oppenheim1981} represents image features in such format which should be high in information and low in redundancy using Fourier transform, rather than set of edges (sharp changes in intensity). in other words, Phase Congruency \cite{Zhang2012} is a dimensionless measure for the of a image structure independently of the signal amplitude which is based on Kovesi's work \cite{Kovesi1999}. Those features are better than gradient-based features which are fully invariant to image illumination and contrast, and also partially invariant  to scale and rotation transformation in case of application of suitable normalization process in frequency domain \cite{Burlacu2008}.

\subsection{Image Normalization}
\begin{figure}
	\centering
		\includegraphics[width=1.00\textwidth]{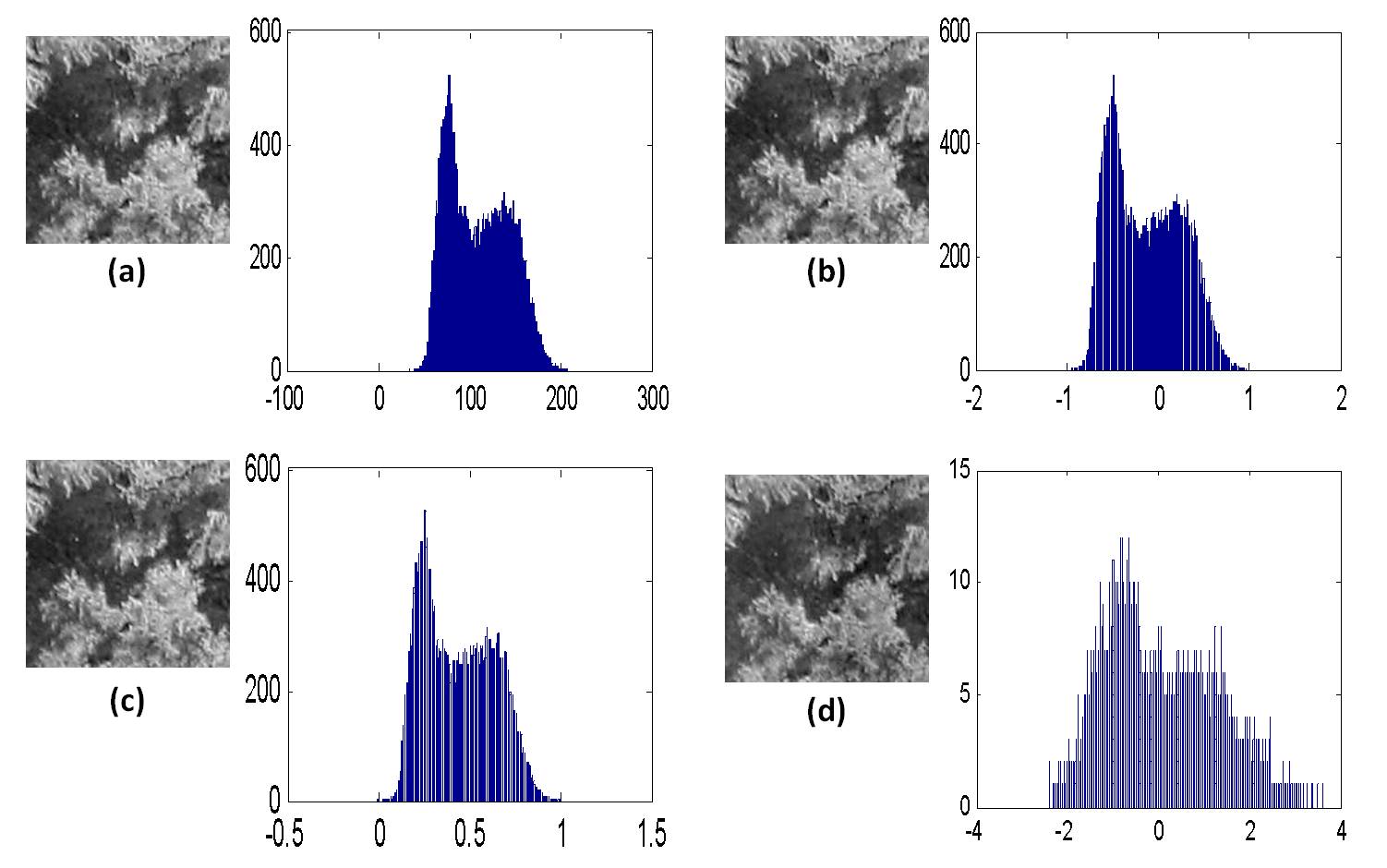}
		\captionsetup{justification=centering,margin=2cm}
	\caption{Examples of image normalization (image, histogram): (a) gray-scale version of original image, (b) min-max normalization [-1,+1], (c) min-max normalization [0,1], (d) z-score normalization }
	\label{fig:Img_Norm}
\end{figure}

There are two different methods for image normalization \cite{Priddy2005} (see figure~\ref{fig:Img_Norm}): linear contrast (min-max) normalization and z-score normalization. it keeps input image within same range as preprocessing step to speed up the network training time.

Z-score normalization (statistical normalization) uses the mean and standard deviation of each input image to normalize its values using linear transformation in order to reduce the effect of outliers (peaky noises) inside input images.
\begin{equation}
\label{eq:z_score}
y=\left[\frac{\left(x-\bar{x}\right)}{s}\right].
\end{equation}
where $\bar{x}$ and $s$ are the mean and standard deviation of the input image $x$ to get an output image $y$ with zero mean and an unit variance.

Min-max normalization (linear contrast normalization) rescales the range of each input image into an unit range [0,1] or a small symmetric range [-1,+1] using linear interpolation formula in order to keep all input images in same scale and allow neural network determining important image features without changing the relationship between image pixels.
\begin{equation}
\label{eq:min_max}
y=(max_{o}-min_{o})*\left[\frac{\left(x-min_{i}\right)}{\left(max_{i}-min_{i}\right)}\right]+min_{o}.
\end{equation}
where $min_{i}$ \& $max_{i}$ are minimum and maximum values of input $x$ range, and $min_{o}$ \& $max_{o}$ are minimum and maximum values of output $y$ range. 

\subsection{Network Architecture}

\subsubsection{Kernel weights \& bias initialization}
The network \cite{LeCun98} initialized biases to zero, and kernel weights using uniform random distribution using the following range:
\begin{equation}
\begin{array} {rcr}
rng = \pm \sqrt{6/(f_{in}+f_{out})}.\\
f_{in} = N_{in}*K^{2}.\\
f_{out} = N_{out}*K^{2}.
\end{array}
\end{equation}
where $N_{in}$ and $N_{out}$ represent number of input and output maps for each hidden layer (i.e. number of input map for layer 1 is 1 as gray-scale image or 3 as color image), and $k$ symbolizes size of convolution kernel for each hidden layer.

\subsubsection{Convolution layer}
Convolution layer construct output maps by convoluting trainable kernel over input maps to extract/combine features for better network behavior using the following equation:
\begin{equation}
\label{eq:conv_layer}
x_{j}^{l}=f(\underset{i \epsilon m_{j}} \sum \left [ x{_{i}}^{l-1} * k_{ij}^{l} + b_{j}^{l} \right ]).
\end{equation}
where $x{_{i}}^{l-1}$ \& $x_{j}^{l}$ are output maps of previous $l-1$ \& current $l$ layers with convolution kernel numbers (input $i$ and output $j$) with weight $k_{ij}^{l}$, $f(.)$ is activation function for calculated maps after summation, and $b_{j}^{l}$ is addition bias of current layer $l$ with output convolution kernel number $j$.

\subsubsection{Down-sampling layer}
The functionality of down-sampling layer is dimensional reduction for feature maps through network's layers starting from input image ending to sufficient small feature representation leading to fast network computation in matrix calculation, which uses the following equation: 
\begin{equation}
\label{eq:ds_layer}
y_{j}^{l} = h_{n}(w^{l}*x_{j}^{l}).
\end{equation}
where $h_{n}$ is non-overlapping averaging function with size $n$x$n$ with neighborhood weights $w$ and applied on convoluted map $x$ of kernel number $j$ at layer $l$ to get less-dimensional output map $y$ of kernel number $j$ at layer $l$ (i.e. 64x64 input map will be reduced using n=2 to 32x32 output map).

\subsubsection{Activation function}
The logistic (sigmoid) function which is the most common activation function for classical neural networks and very useful in gradient decent training due to existence of function's derivatives. the function's equation is as follows:
\begin{equation}
f(x) = \frac{1}{1+e^{-\beta x}} ; [-\infty,+\infty] \Rightarrow [0,1].
\end{equation}
where input $x$ can be infinite value, and output $f(x)$ will be in bounded range [0,1].

\subsubsection{Learning rate}
Inspired from Lawrence's convergence learning rate in CNN application for face recognition \cite{Lawrence1997}, an adapt learning rate is used rather than a constant one with respect to network's status and performance as follows: 
\begin{equation}
\label{eq:lr}
\alpha^{n} = g(\frac{\alpha ^{n-1}}{\left [ n / \left ( N/2 \right ) \right ]+1}+e^{n}).
\end{equation}
where $\alpha^{n}$ \& $\alpha^{n-1}$ are learning rates of current \& previous iterations (if first network iteration is the current one, then learning rate of previous network iteration represents initial learning rate as network input), $n$ \& $N$ are number of current network iteration \& total number of iterations, $e^{n}$ is back-propagated error of current network iteration, and $g(.)$ is linear limitation function to keep value of learning rate in range $(0,1]$.

\subsubsection{Error back-propagation}
The network is back-propagated with squared-error loss function as follows:
\begin{equation}
E^{n} = \frac{1}{2} \sum_{n=1}^{N}\sum_{k=1}^{C}(t_{k}^{n}-y_{k}^{n})^{2}.
\end{equation}
where $N$ \& $C$ are number of training samples \& output classes, and $t$ \& $y$ are target \& actual outputs 

\section{Summary}
This chapter showed the proposed framework for coral reef classification using deep learning and explained in details how implementation process from color enhancement and data adjustment as preprocessing step to feature extraction and classification using convolutional neural networks.

\chapter{Results} \label{chap:results}
This chapter shows the results of sparse classification with hybrid patching around annotated points using convolutional neural networks initially referring to Palm's toolbox for deep learning \cite{Palm2012}, in which it discusses introduction of new coral dataset ADS besides an existed dataset MLC from University of California San Diego, explanation of used evaluation metrics, experimental and final results for best configuration selection, and finally output representation of the proposed method with respect to selected configuration. 

\section{Datasets}

\subsection{Moorea Labeled Corals}

\begin{figure}
	\centering
		\includegraphics[width=1.00\textwidth]{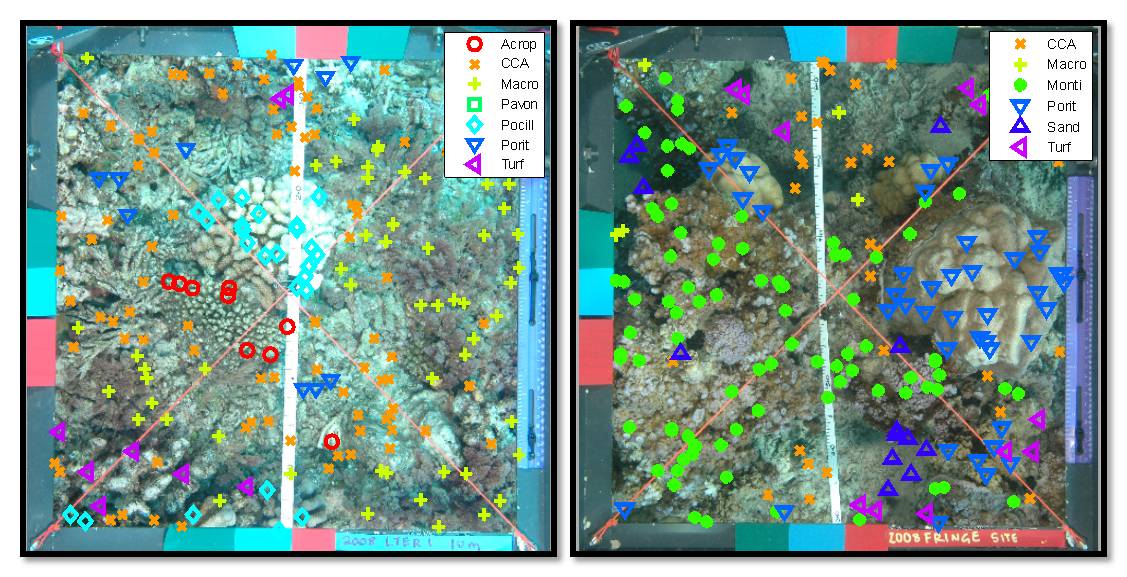}
	\caption{Sample images from UCSD's MLC dataset}
	\label{fig:MLC}
\end{figure}

University of California, San Diego (UCSD)'s Moorea labeled corals ``MLC" \cite{Beijbom2012} dataset is captured from the island of Moorea in French Polynesia in which point-based annotations are provided (200 points per image) in around two-thousand images categorized in three different years (2008,2009,2010). The concerned labels form 9 coral/non-coral classes (as shown in figure~\ref{fig:MLC}), These classes are classified into 5 coral classes (Acropora ``Acrop" , Pavona ``Pavon" , Montipora ``Monti" , Pocillopora ``Pocill" , and Porites ``Porit") and 4 non-coral classes (Crustose Coralline Algae ``CCA" , Turf algae ``Turf" , Macroalgae ``Macro" , and Sand ``Sand").

\subsection{Atlantic Deep Sea}

\begin{figure}
	\centering
		\includegraphics[width=0.5\textwidth]{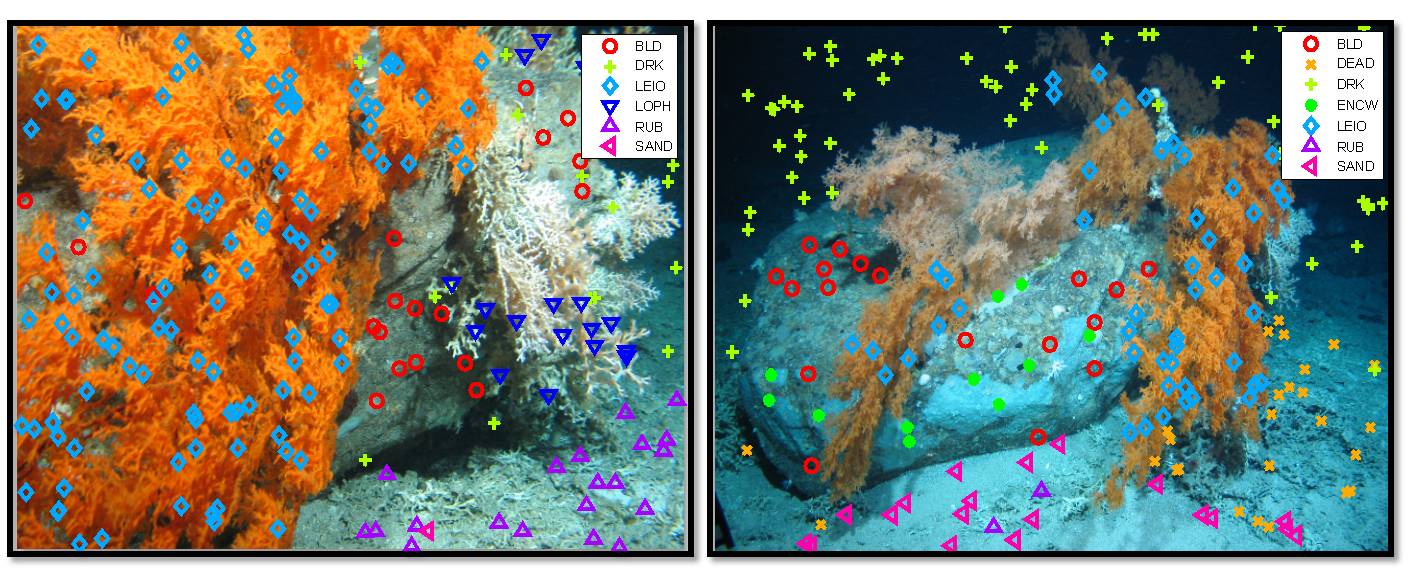}
	\caption{Sample images from HWU's ADS dataset}
	\label{fig:ADS}
\end{figure}

Heriot-Watt University (HWU)'s Atlantic Deep Sea (ADS) dataset \cite{Roberts2013} represents cold-water coral reefs from north Atlantic west of Scotland and Ireland in 2012 at Depth (100-800) meters. around 50 images are expertly annotated (200 labeled points per image) clarifying different types of Lophelia coral habitats and the surrounding soft sediment Logachev mounds (Rockall Trough). The target nine classes are classified (as shown in figure~\ref{fig:ADS}) into 5 coral classes (DEAD ``Dead Coral" , ENCW ``Encrusting White Sponge" , LEIO ``Leiopathes Species" , LOPH ``Lophelia" , and RUB ``Rubble Coral") and 4 non-coral classes (BLD ``Boulder" , DRK ``Darkness" , GRAV ``Gravel" , and Sand ``Sand").

\section{Evaluation Metrics}
There are many popular assessment methods for quantitative measures in classification problems. The statistics of confusion matrix (contingency matrix)\cite{Couto2003} is general quantitative representation of relationship between target classes and algorithm output classes, resulting of some important accuracy quantities (overall accuracy ``OA", precision, recall, sensitivity, specificity, and F-score). training and test errors are also used to validate classification performance over different selection of network parameters.

\section{Experimental Results}

\subsection{Network parameters}
Finding best network architecture and validating its performance needs to compare quantitative results with keeping the rest network parameters constant (size of hybrid input image = 181x181, number of output classes = 9, number of samples per class = 300, number of input channels = 3 as RGB image, normalization method = min-max with in range [-1,+1], initial learning rate = 1, network batch size = 3, number of network epochs = 10, number of hidden output maps = 6-12, and ratio of training/test sets = 2:1).

\subsection{Color enhancement}

\begin{figure}
\centering
\begin{subfigure}[b][][b]{.5\textwidth}
  \centering
  \includegraphics[width=.9\linewidth]{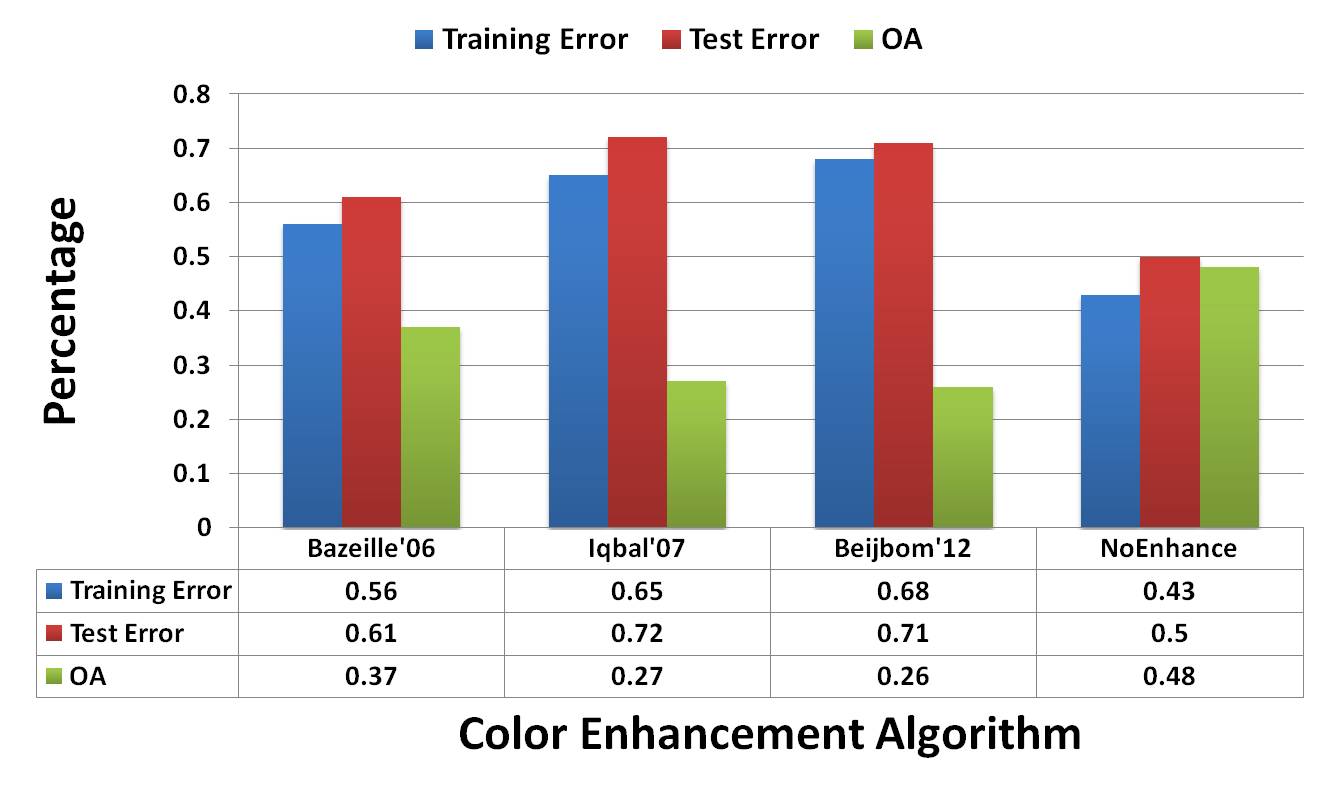}
  \caption{MLC dataset}
  \label{fig:MLC_Color}
\end{subfigure}%
\begin{subfigure}[b][][b]{.5\textwidth}
  \centering
  \includegraphics[width=.9\linewidth]{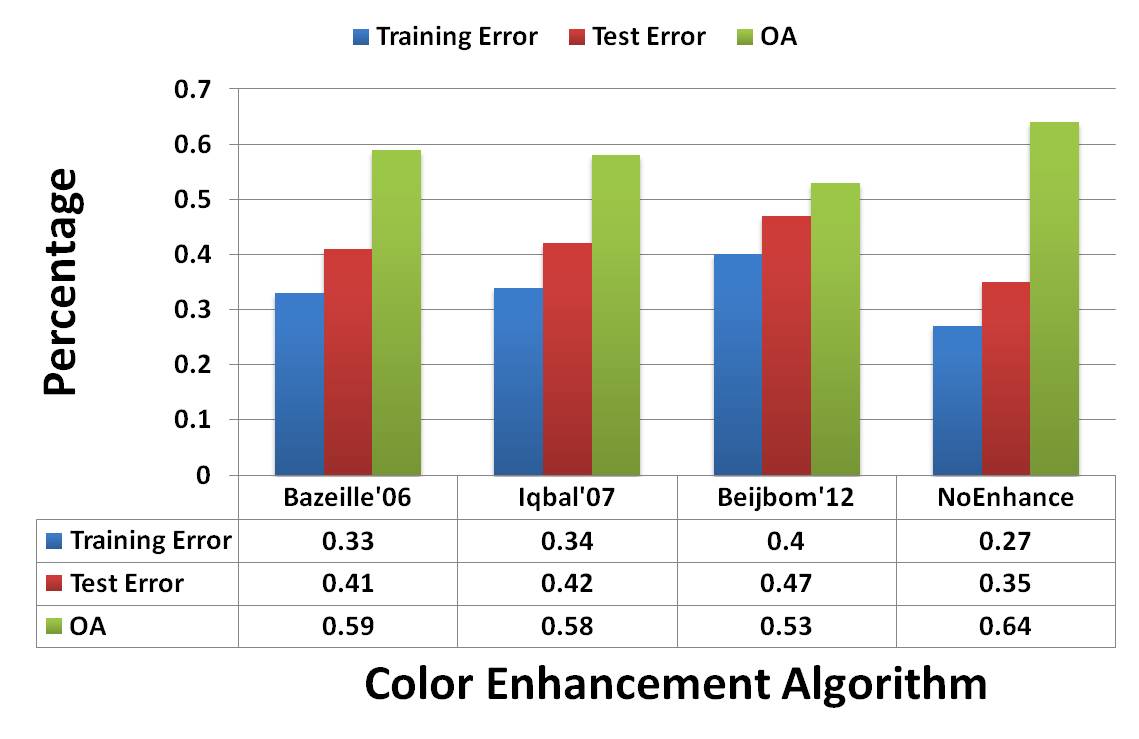}
  \caption{ADS dataset}
  \label{fig:ADS_Color}
\end{subfigure}
\caption{Color enhancement comparison}
\label{fig:Color}
\end{figure}

Figure~\ref{fig:MLC_Color} shows in MLC dataset that Bazeille'06 is the best color enhancement algorithm in classification results (around 10\% improvement in all quantities: training error, test error, and overall accuracy) over other algorithms (Iqbal'07, Beijbom'12). Although raw image data without any enhancement is the best preprocessing choice for network classification within more than 10\% value difference from nearest enhancement algorithm. Figure~\ref{fig:ADS_Color} using ADS dataset confirms the stated results in MLC dataset, but with small difference in values.

%\begin{figure}
	%\centering
		%\includegraphics[width=1.00\textwidth]{figures/Results/MLC_Color.jpg}
	%\caption{Color enhancement comparison using MLC dataset}
	%\label{fig:MLC_Color}
%\end{figure}

\subsection{Hybrid patching}

\begin{figure}
\centering
\begin{subfigure}[b][][b]{.5\textwidth}
  \centering
  \includegraphics[width=.9\linewidth]{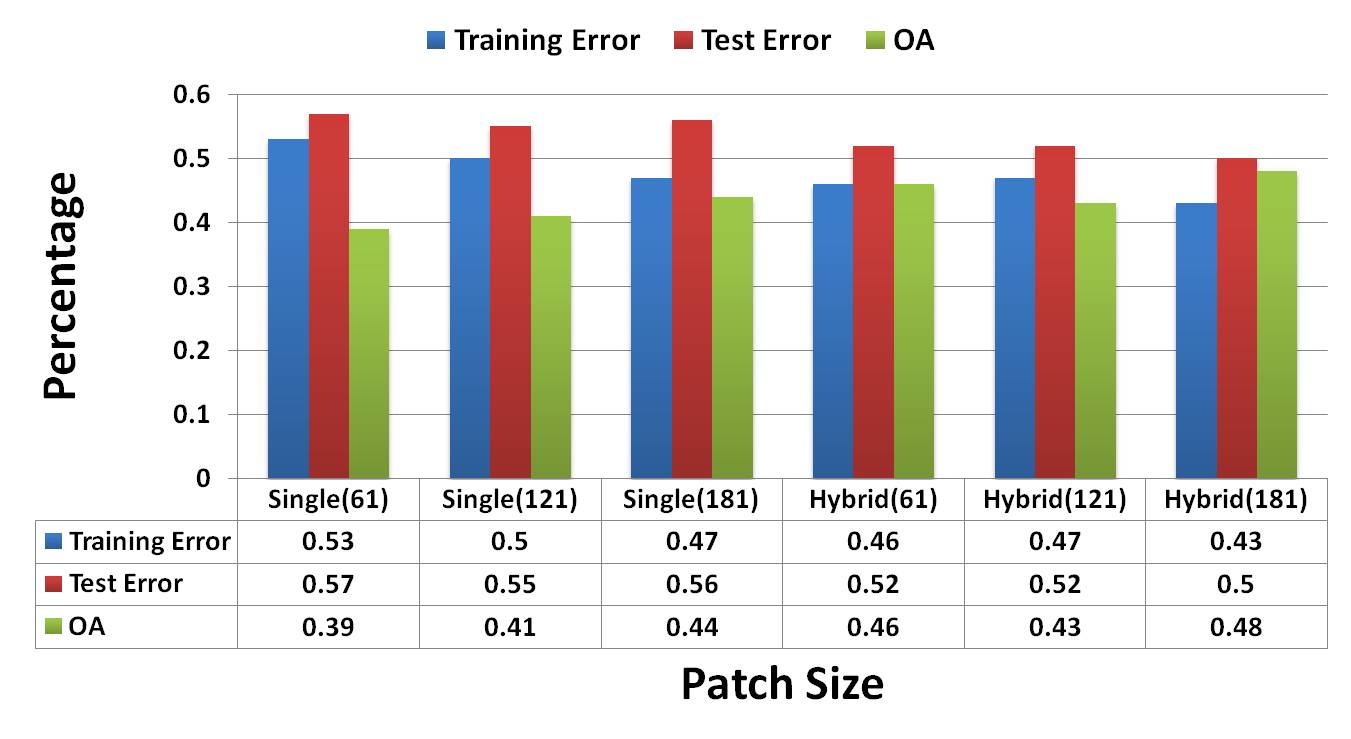}
  \caption{MLC dataset}
  \label{fig:MLC_Patch}
\end{subfigure}%
\begin{subfigure}[b][][b]{.5\textwidth}
  \centering
  \includegraphics[width=.9\linewidth]{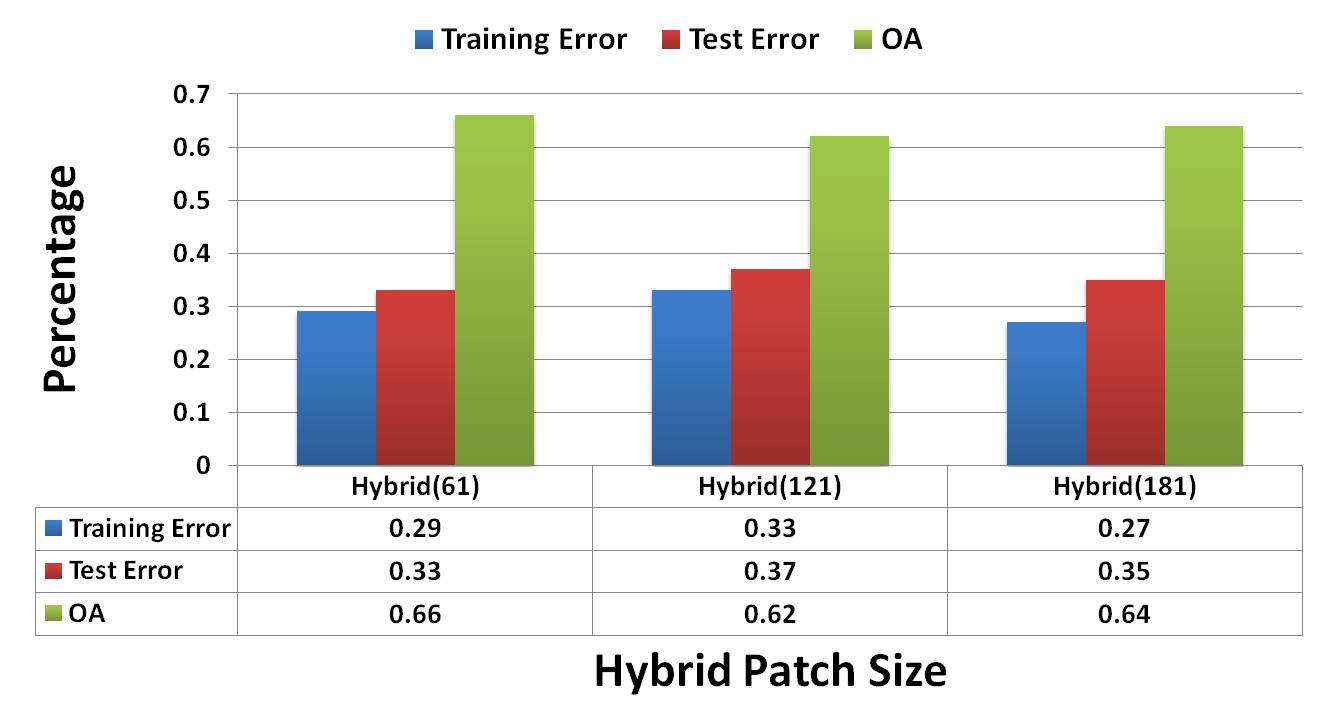}
  \caption{ADS dataset}
  \label{fig:ADS_Patch}
\end{subfigure}
\caption{Patch selection comparison}
\label{fig:Patch}
\end{figure}

Figure~\ref{fig:MLC_Patch} using MLC dataset presents less error rates in unified-scaling multi-size image patches over single-sized image patches,and  up-scaling in multi-size image patches have the best comparison results across different measurements (least training and test errors and highest overall accuracy value) with small difference in overall accuracy value 2\% from hybrid down-scaling. Due to insufficient number of patch samples, ADS dataset in figure~\ref{fig:ADS_Patch} does only hybrid comparison only, and it states the opposite results saying that hybrid down-scaling has the better performance against hybrid up-scaling with the same difference in accuracy value of MLC dataset. Bi-cubic interpolation is used in hybrid patching (built-in MATLAB function ``imresize''). For a larger image size, classification performance has opposite proportion with computation time in most cases. Hybrid down-scaling (61) is finally selected for large-scale experiments.

\subsection{Feature maps}
\begin{figure}
\centering
\begin{subfigure}[b][][b]{.5\textwidth}
  \centering
  \includegraphics[width=.9\linewidth]{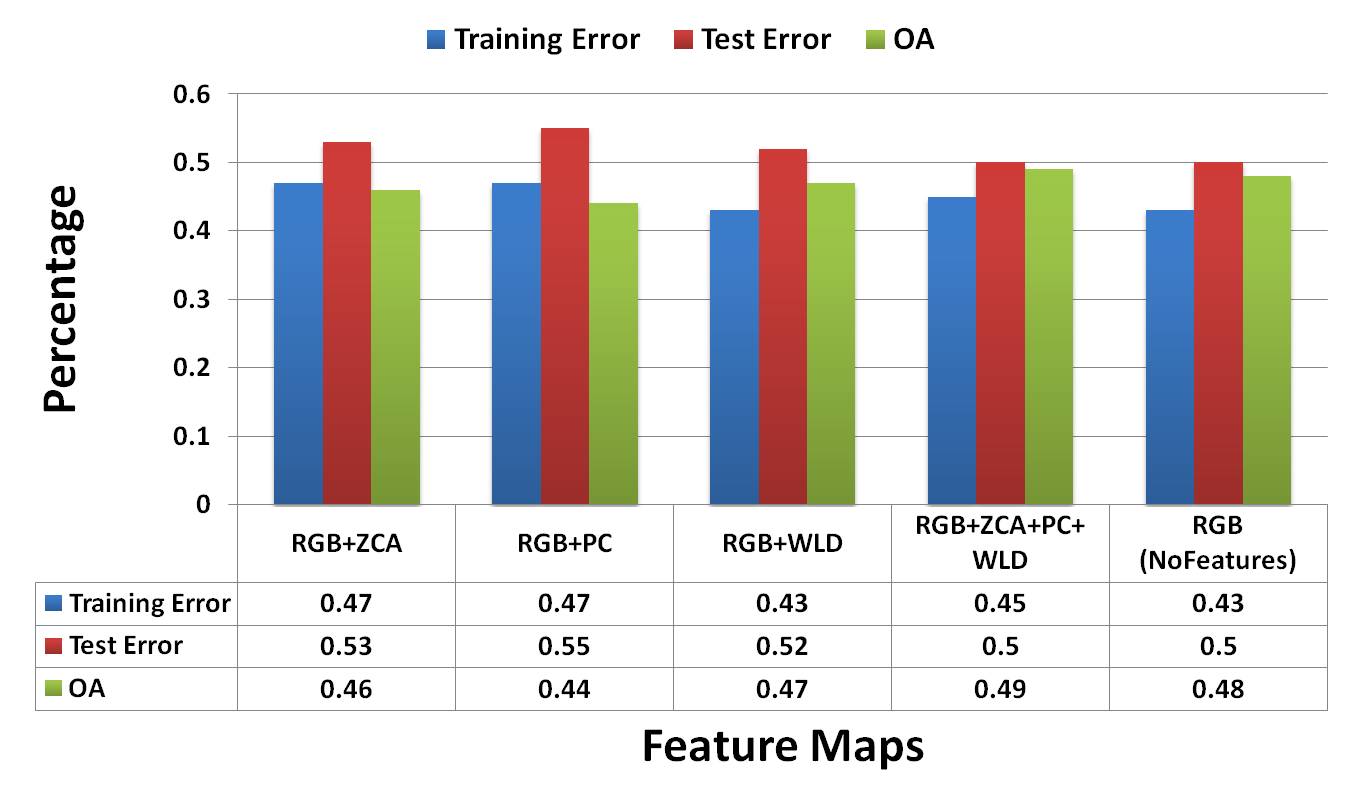}
  \caption{MLC dataset}
  \label{fig:MLC_Feature}
\end{subfigure}%
\begin{subfigure}[b][][b]{.5\textwidth}
  \centering
  \includegraphics[width=.9\linewidth]{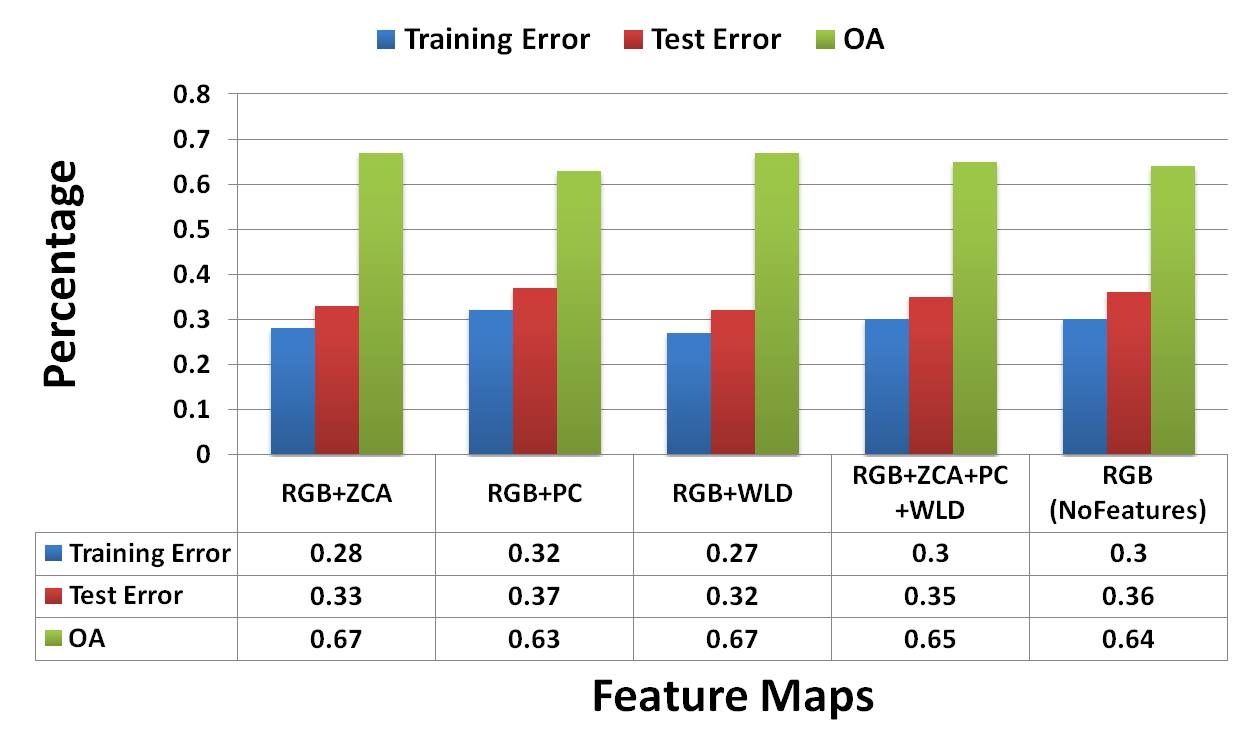}
  \caption{ADS dataset}
  \label{fig:ADS_Feature}
\end{subfigure}
\caption{Feature maps comparison}
\label{fig:Feature}
\end{figure}

Figure~\ref{fig:Feature} indicates that additional feature-based channel besides basic color channels will be useful in coral discrimination in both datasets (MLC,ADS). Such that combination of three feature-based maps has slightly better classification results (1\% difference in overall accuracy) in both datasets over basic color channels without any additional supplementary channels. More large-scale experiments are needed to decide in more discriminative way if additional feature-based channels will improve classification performance or not.

\subsection{Image normalization}
\begin{figure}
\centering
\begin{subfigure}[b][][b]{.5\textwidth}
  \centering
  \includegraphics[width=.9\linewidth]{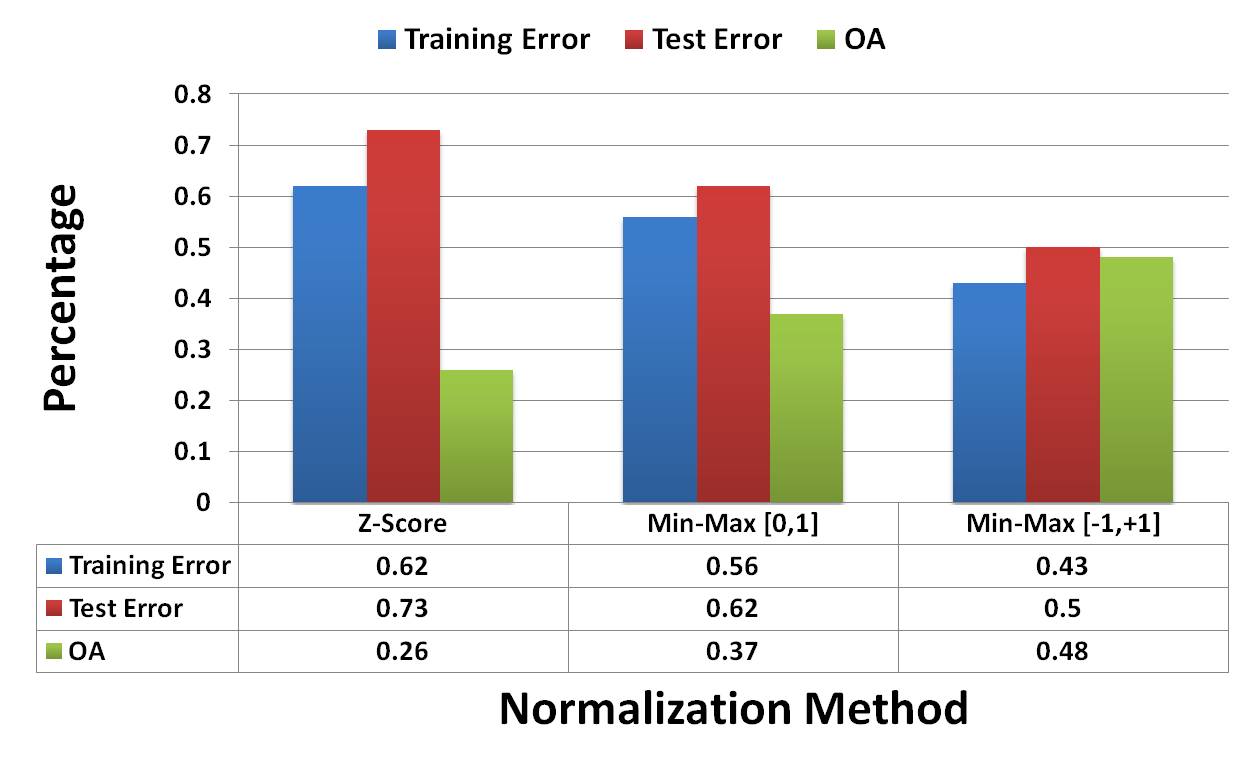}
  \caption{MLC dataset}
  \label{fig:MLC_Normalization}
\end{subfigure}%
\begin{subfigure}[b][][b]{.5\textwidth}
  \centering
  \includegraphics[width=.9\linewidth]{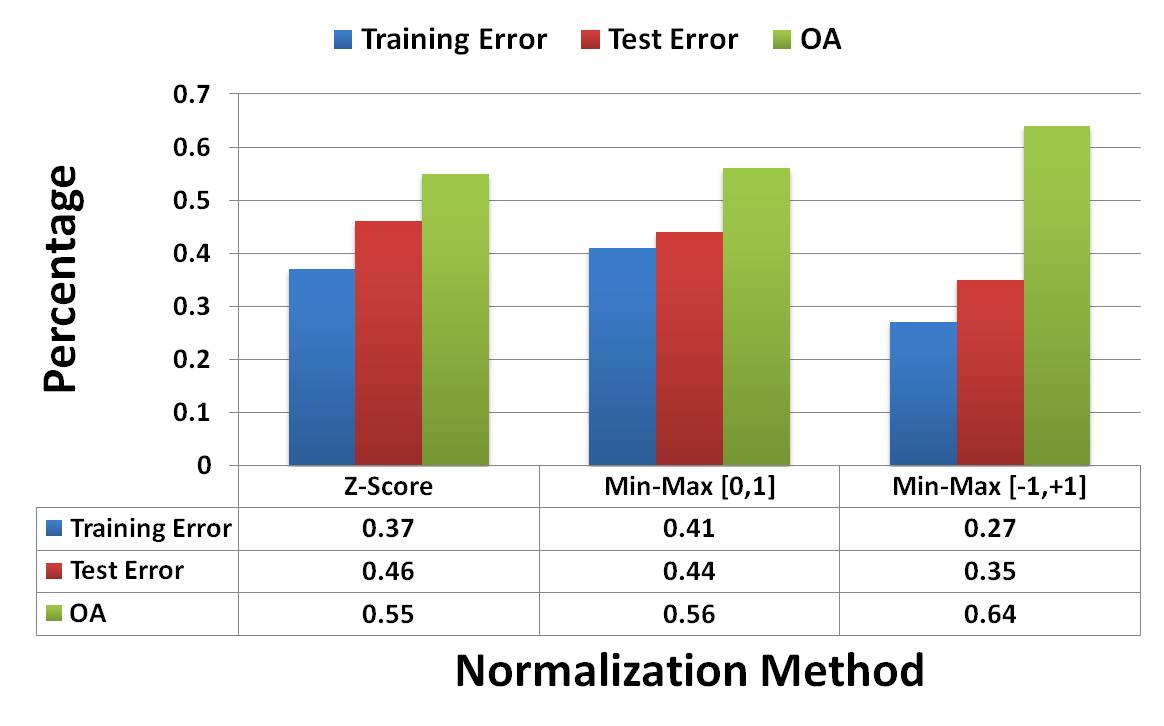}
  \caption{ADS dataset}
  \label{fig:ADS_Normalization}
\end{subfigure}
\caption{Normalization methods comparison}
\label{fig:Normalization}
\end{figure}

in MLC \& ADS datasets in figure~\ref{fig:Normalization}, z-score normalization has very bad classification results in comparing with min-max normalization such that changing data to be zero-mean values and unit-variance affects negatively in classification process. Applying more range in min-max normalization has very positive classification impact, in such that min-max with range [-1,+1] will be selected for further large-scale experiments.

\subsection{Hidden output maps}
\begin{figure}
\centering
\begin{subfigure}[b][][b]{.5\textwidth}
  \centering
  \includegraphics[width=.9\linewidth]{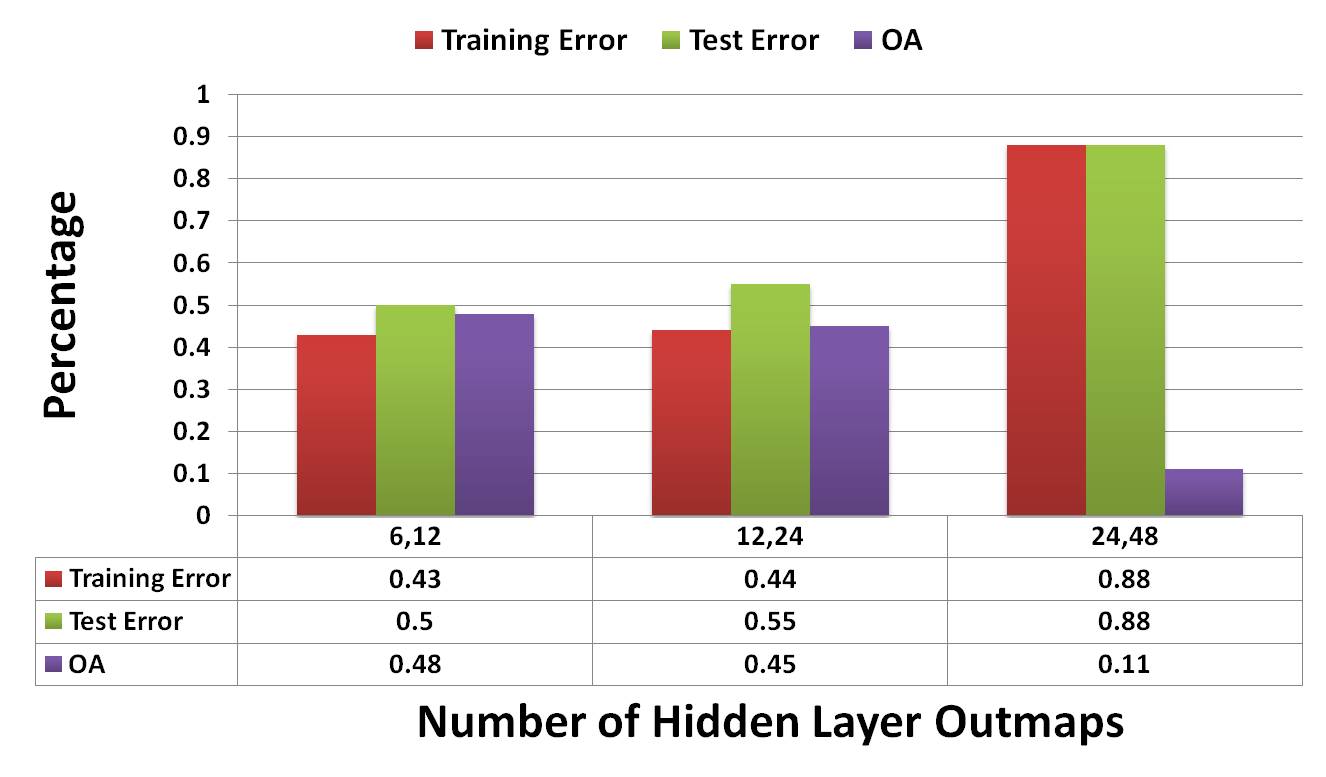}
  \caption{MLC dataset}
  \label{fig:MLC_Outmaps}
\end{subfigure}%
\begin{subfigure}[b][][b]{.5\textwidth}
  \centering
  \includegraphics[width=.9\linewidth]{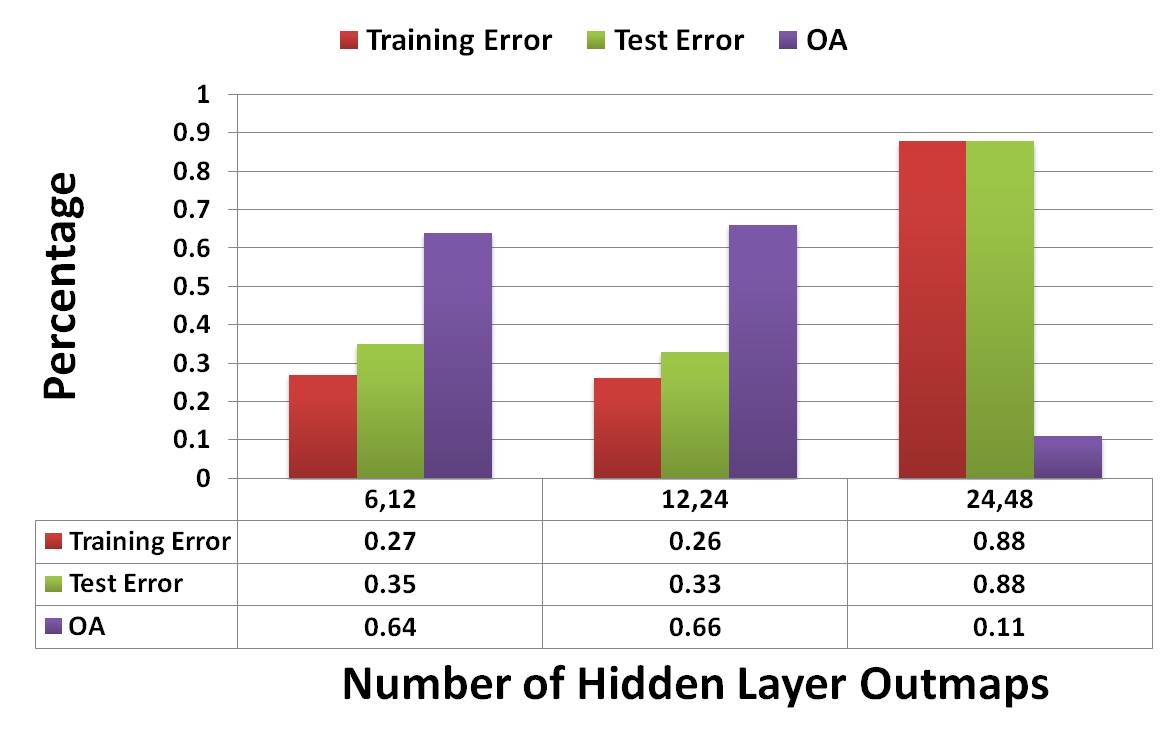}
  \caption{ADS dataset}
  \label{fig:ADS_Outmaps}
\end{subfigure}
\caption{Capacity comparison of hidden output maps}
\label{fig:Outmaps}
\end{figure}

As seen in figure~\ref{fig:Outmaps}, using outrageous number (24-48) of hidden output maps makes classification algorithm behaving inappropriately leading to converge all input data to only one output class. The presented classification results can't indicates the most suitable number for hidden output maps
between (6-12) and (12-24).
 
\subsection{Summary}
After experimenting different classification configurations, down-scaled hybrid RGB input image are normalized using min-min method with range [-1,+1] and selected with additional optional feature-based maps as input data for convolutional neural networks with one of two different number of hidden output maps (6-12 or 12-24).

\section{Final Results}
\begin{figure}
\centering
\begin{subfigure}[b][][b]{.5\textwidth}
  \centering
  \includegraphics[width=.9\linewidth]{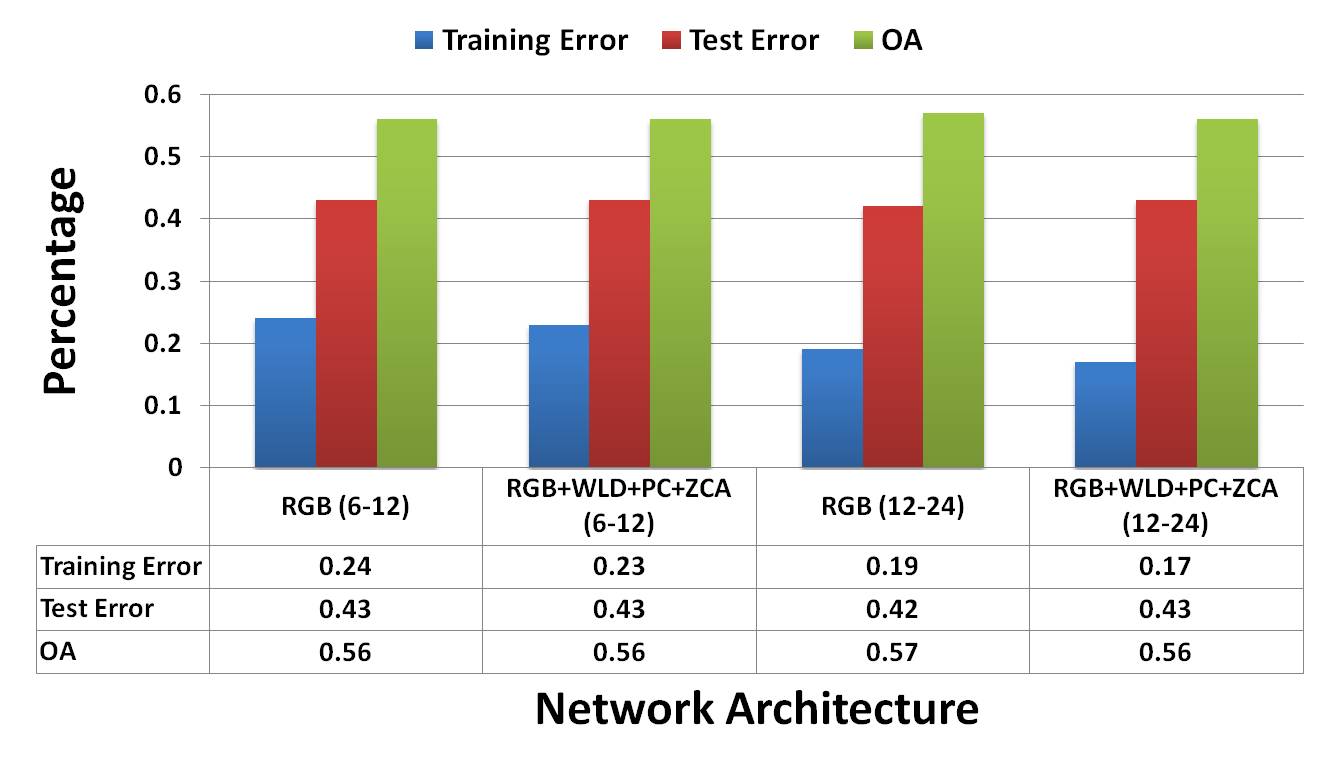}
  \caption{MLC dataset}
  \label{fig:MLC_50}
\end{subfigure}%
\begin{subfigure}[b][][b]{.5\textwidth}
  \centering
  \includegraphics[width=.9\linewidth]{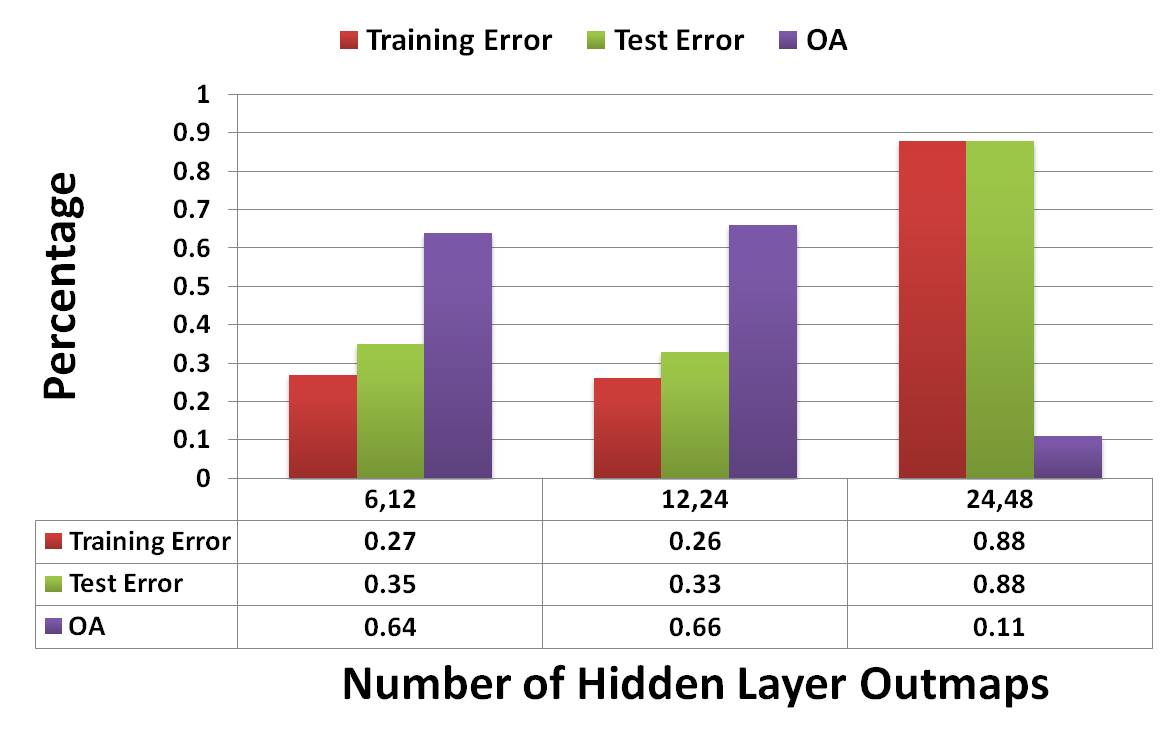}
  \caption{ADS dataset}
  \label{fig:ADS_50}
\end{subfigure}
\caption{Comparison of network architecture}
\label{fig:Net50}
\end{figure}

In large-scale experiments (50 epochs rather than 10 epochs), testing phase of MLC dataset has almost the same results as shown in figure~\ref{fig:MLC_50}, but training phase starts converging to correct target classes by increasing number of hidden output maps (12-24) and using additional feature-based maps as supplementary channels. Using ADS dataset in figure~\ref{fig:ADS_50}, testing phase has best significant accuracy results with same selected configuration for MLC dataset.

\begin{figure}
\begin{subfigure}{.5\linewidth}
\centering
\includegraphics[width=.9\linewidth]{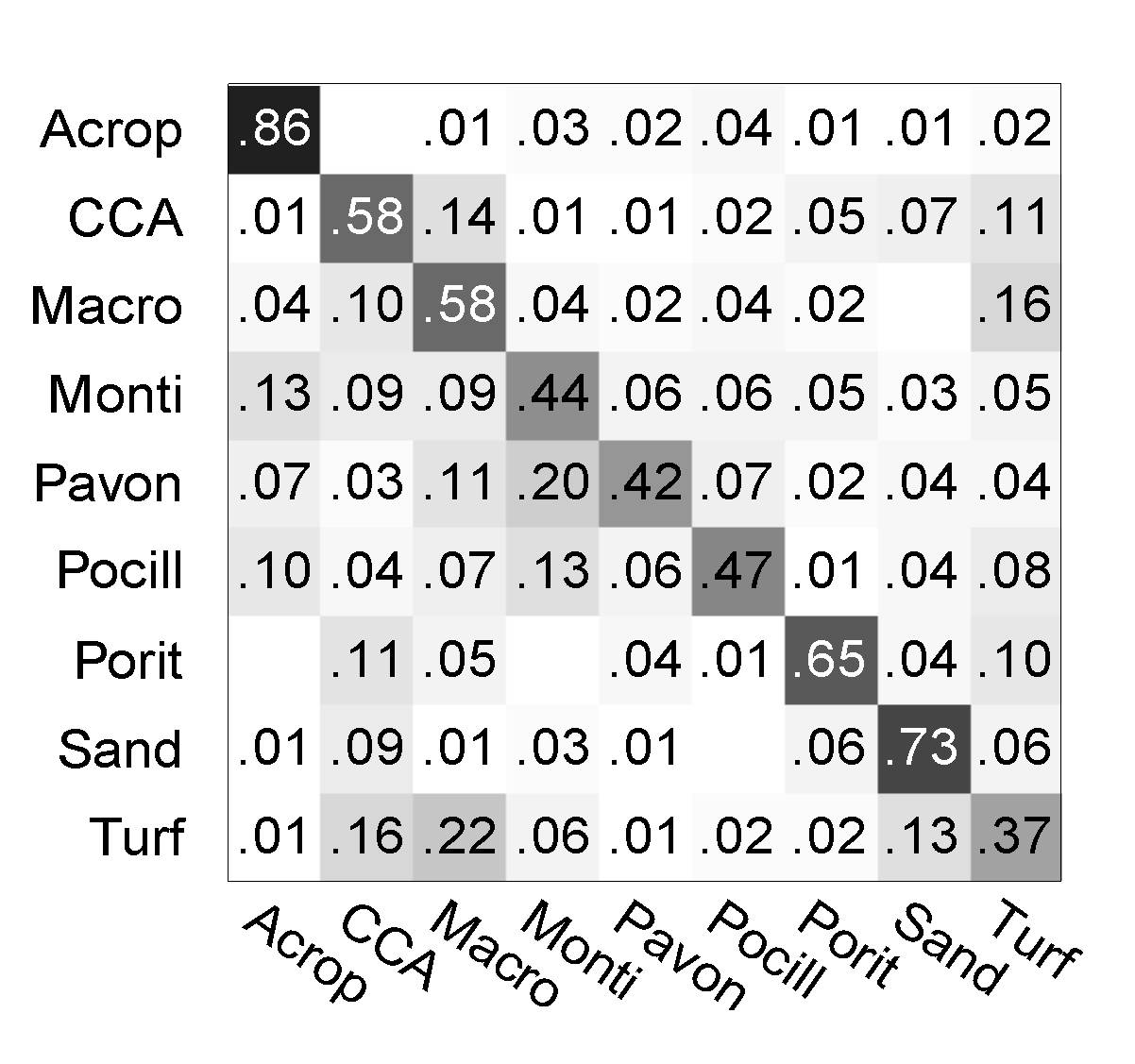}
\caption{}
\label{fig:MLC_CM}
\end{subfigure}%
\begin{subfigure}{.5\linewidth}
\centering
\includegraphics[width=.9\linewidth]{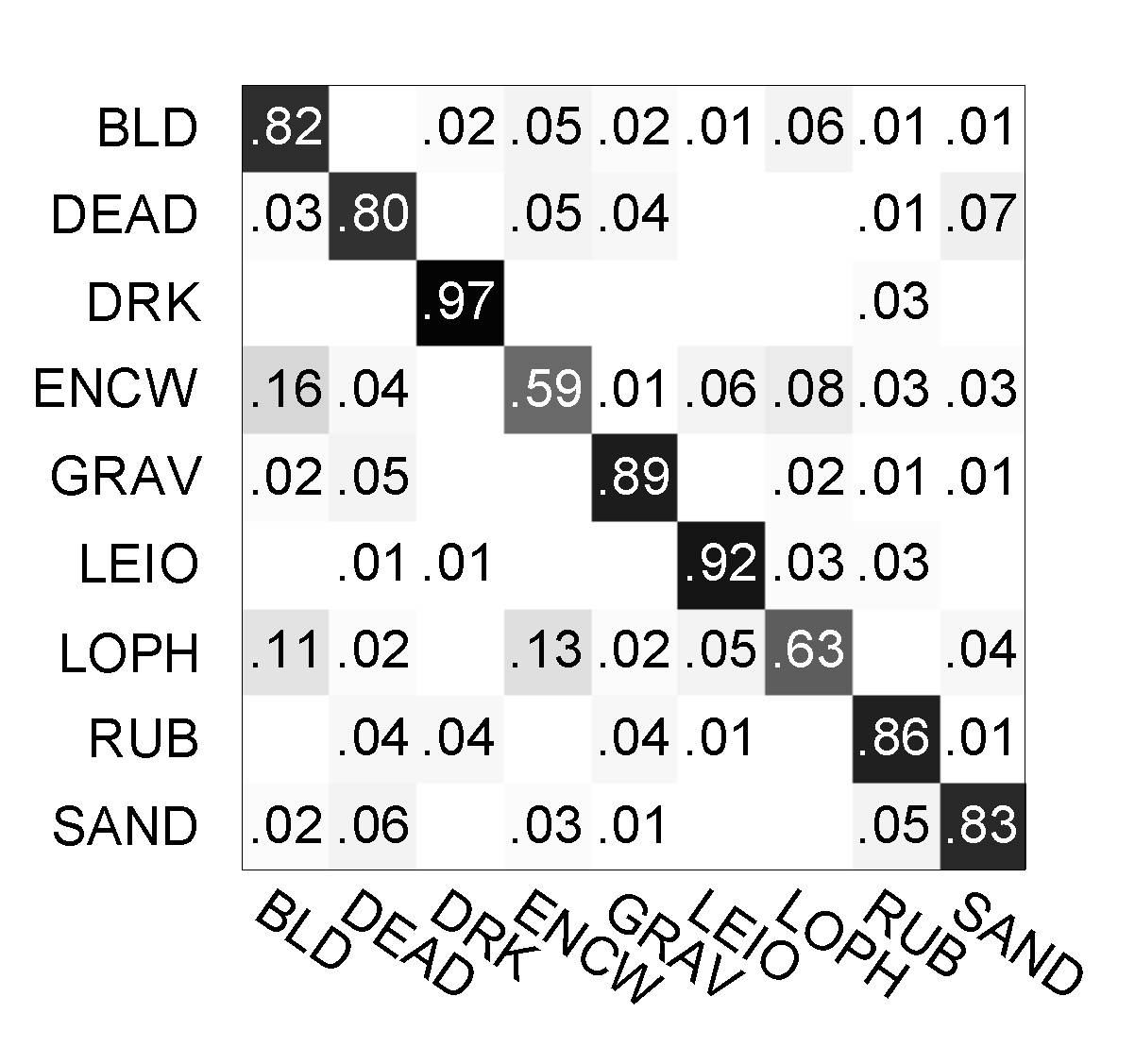}
\caption{}
\label{fig:ADS_CM}
\end{subfigure}
\begin{subfigure}{.5\linewidth}
\centering
\includegraphics[width=.9\linewidth]{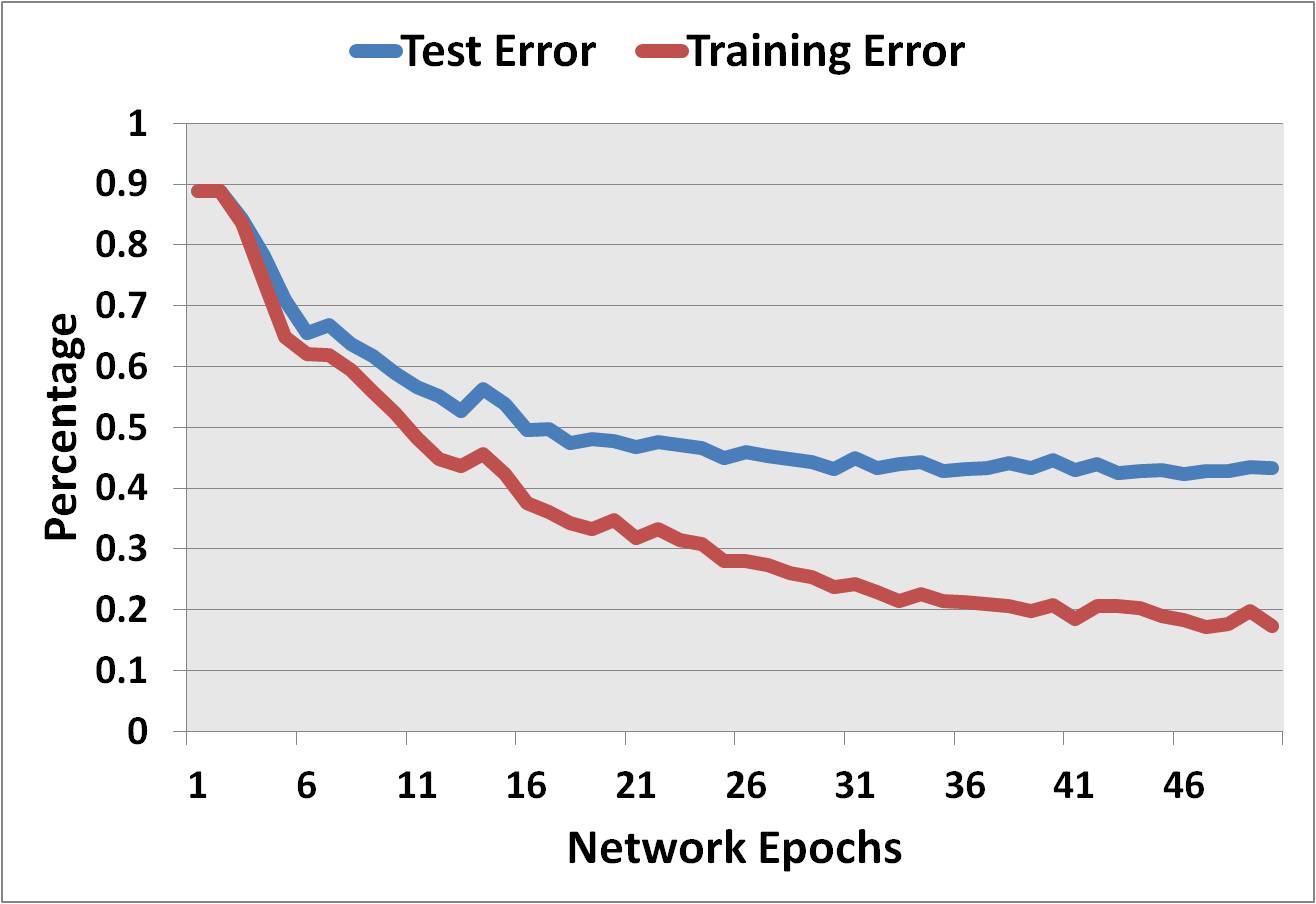}
\caption{}
\label{fig:MLC_ERR}
\end{subfigure}%
\begin{subfigure}{.5\linewidth}
\centering
\includegraphics[width=.9\linewidth]{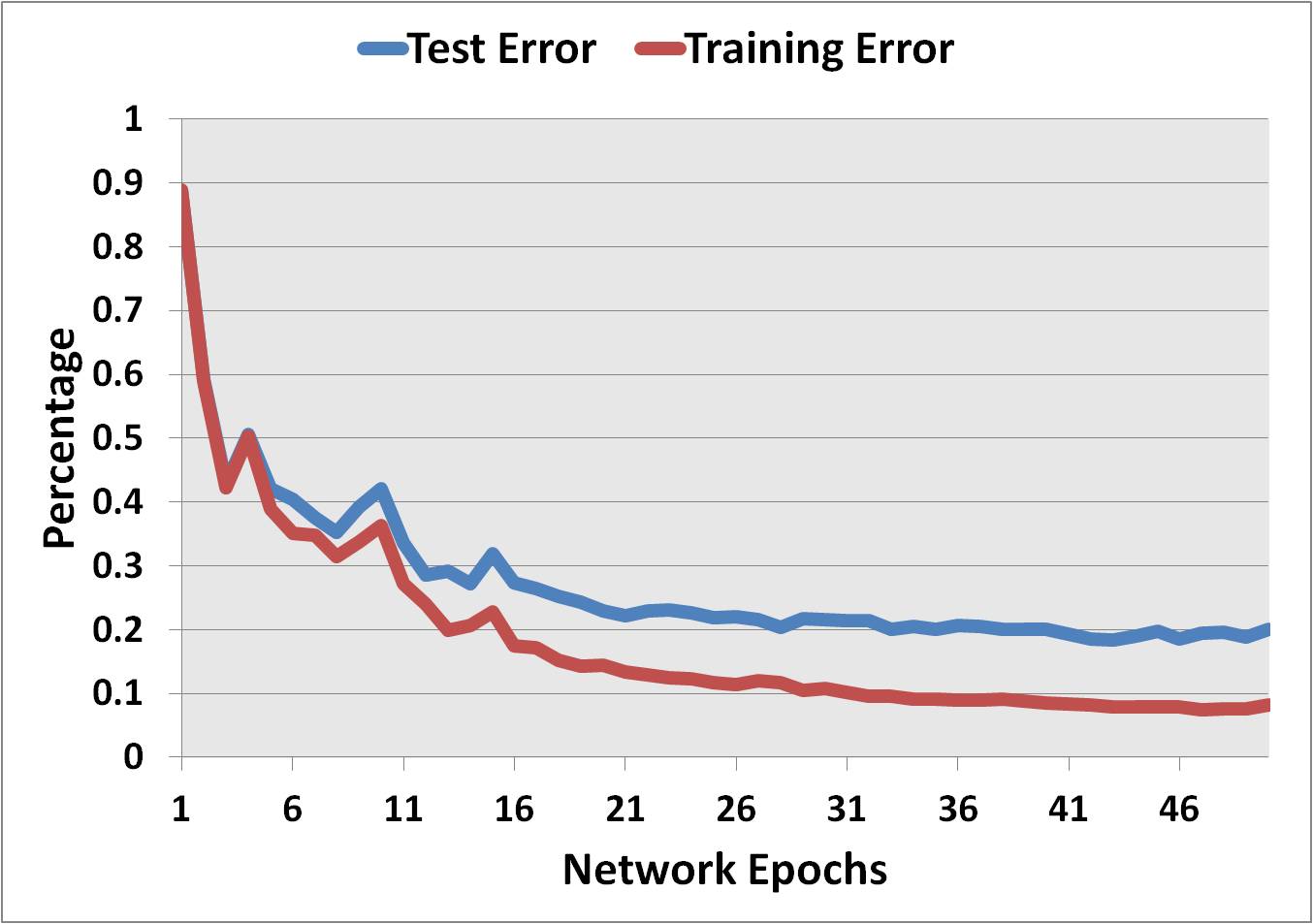}
\caption{}
\label{fig:ADS_ERR}
\end{subfigure}
\begin{subfigure}{.5\linewidth}
\centering
\includegraphics[width=.9\linewidth]{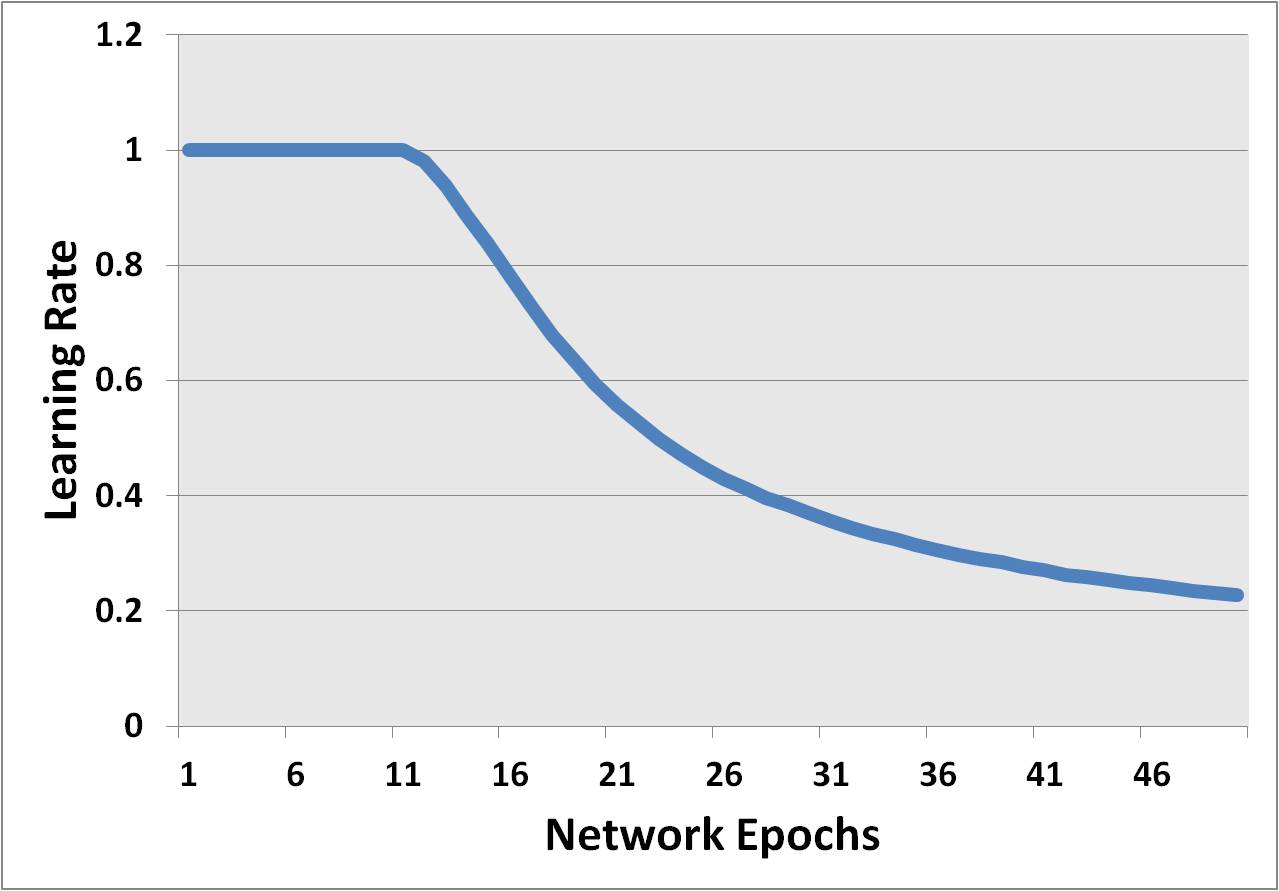}
\caption{}
\label{fig:MLC_LR}
\end{subfigure}%
\begin{subfigure}{.5\linewidth}
\centering
\includegraphics[width=.9\linewidth]{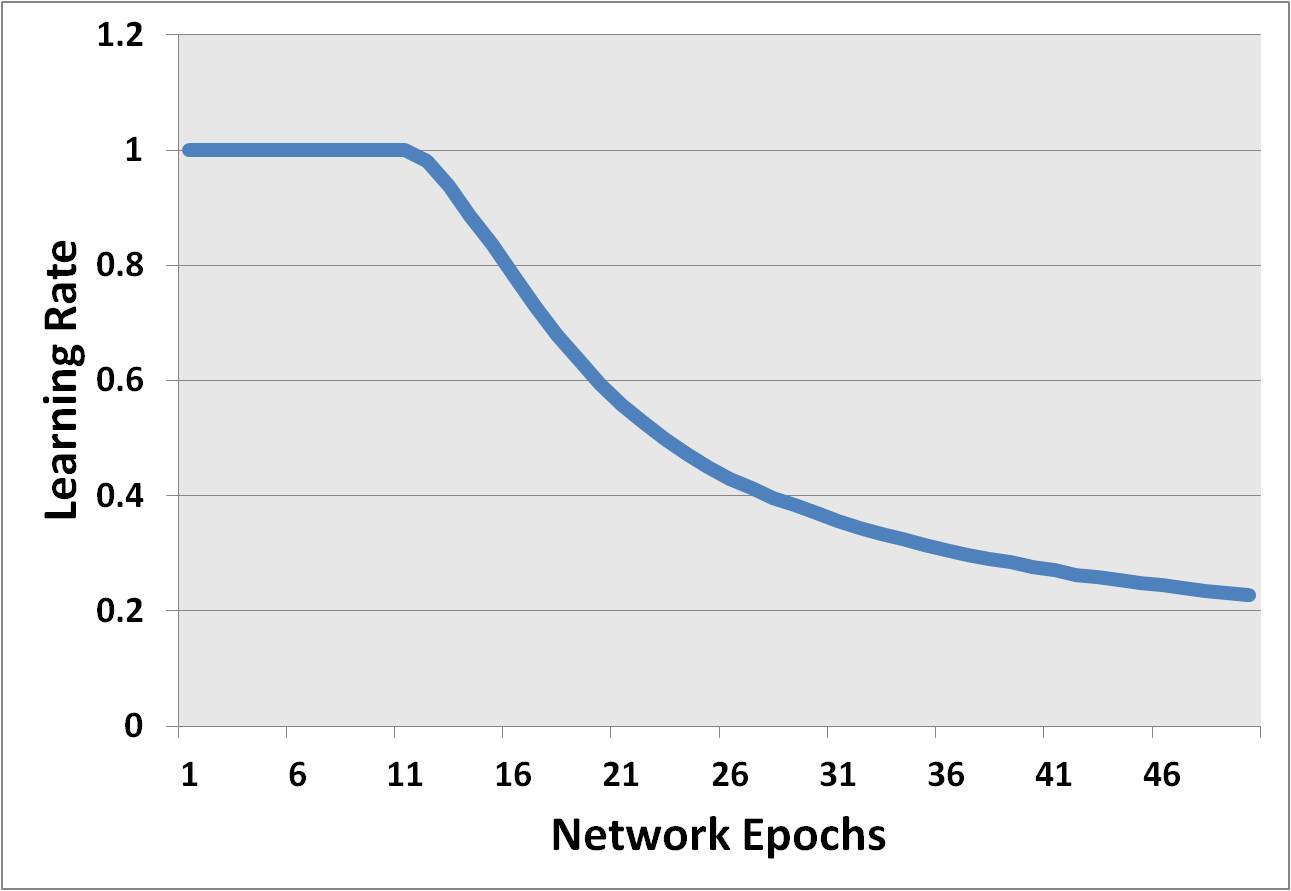}
\caption{}
\label{fig:ADS_LR}
\end{subfigure}
\captionsetup{justification=centering,margin=2cm}
\caption{Evaluation Metrics for selected network architecture: (a,c,e) MLC dataset, (b,d,f) ADS dataset}
\label{fig:test}
\end{figure}

Sub-figures~\ref{fig:MLC_CM},~\ref{fig:ADS_CM} represent confusion matrices for MLC and ADS dataset, in which rows \& columns represent the assignments of target classes \& predicted output classes respectively. In MLC dataset, the highest classification rates are for Acrop (coral) and Sand (non-coral), and the lowest classification rates are for Pavon (coral) and Turf (non-coral), where misclassification occurred outputting Pavon as Monti/Macro and Turf as Macro/CCA/Sand due to similarity in their shape properties or growth environment. However in ADS dataset, non-corals has better classification rate then corals, where DRK (non-coral) has almost perfect classification rate due to its distinct nature (almost dark blue plain image), LEIO (coral) has excellent classification rate due to its distinction color property (orange), and LOPH (coral) \& ENCW (coral) has lowest classification rates due to their color confusion with each other \& with BLD (non-coral).

Sub-figures~\ref{fig:MLC_ERR},~\ref{fig:ADS_ERR} show the evolution of training and test errors in MLC \& and ADS datasets across network epochs, such that the proposed method's errors have better convergence curves (almost half) with ADS dataset over the other one. From epoch 30 in MLC dataset, increased gap starts to appear between training and test errors leading to algorithm over-fitting over training data. From epoch 35 in ADS dataset, training and test errors are almost stagnant (no major improvement) with respect to typical evolution of neural networks. MLC \& ADS datasets have similar evolution curves (Sub-figures~\ref{fig:MLC_LR},~\ref{fig:ADS_LR}) for learning rate across presented network epochs.
\chapter{Conclusions} \label{chap:conclusions}

This chapter summarizes the outcomes of proposed thesis, in which the following sections present an overview of the proposed work, discuss the main contributions, and finally introduce limitations in the implementation of proposed work and future work in the same research direction.

\section{Summary}
This thesis discussed the importance of deep/shallow sea corals, their main threads from human interaction and environmental changes, their human-manual transplantation solution through scuba divers, and the involvement of under-water robots in automatic development technique in coral detection \& transplantation using camera sensors. It covered all recent research work in coral classification using images captured from remotely operated underwater vehicles, and introduction of modern deep supervised classification method (convolutional neural networks) \& its recent applications in related research fields (object detection and classification). Using new introduced deep-sea coral dataset from Heriot-Watt University along side with shallow-sea coral dataset from University of California San Diego, it proposed a supervised sparse-based classification method for coral species using convolutional neural networks as feature extraction \& classification, and investigated computation of supplementary channels (feature-based maps) besides basic spatial color channels (spatial-based maps) to act as coral input data with state-of-art preprocessing underwater algorithms for image enhancement and color normalization. It finally suggested well-defined future research vision in under-water imaging applications using deep learning methods.

\section{Main Contributions}
The proposed framework in this thesis presented some contributions as follows:
\begin{itemize}
\item First application of deep learning techniques (especially convolutional neural networks) in under-water image processing (detection or classification).
\item Introduction of new coral-labeled dataset ``Atlantic Deep Sea" representing cold-water coral reefs from Scotland and Ireland.
\item Investigation of convolutional neural networks in handling noisy large-sized images, manipulating point-based multi-channel input data.
\item Hybrid image patching procedure for multi-size scaling process across different square-based windowing around labeled points.
\item Production of two pending publications in ICPR-CVAUI 2014 (22nd International Conference on Pattern Recognition Workshop: Computer Vision for Analysis of Underwater Imagery), and ACCV 2014 (12th Asian Conference on Computer Vision).
\end{itemize}

\section{Limitations}
The proposed classification framework has the following limitations:
\begin{itemize}
\item Lack of fast performance of proposed algorithm and handling large-sized input data.
\item In-comprehensive assessment comparison against other coral classification methods.
\item Difficulty in finding optimal-fit structure and parameters for deep convolutional neural networks due to insufficient references for new research deep learning techniques.
\item Absence of uniform distribution for labeled coral classes and continuous depth calculation for further scale-based operations.
\end{itemize}

\section{Future Work}
The future work of proposed method will cover avoiding information loss of dimension reduction for convolutional neural networks (no hidden sub-sampling layers will be used), composition of multiple deep convolutional models for N-dimensional data (different learning processes for basic and extra channels of input data), development of real-time image/video application for coral recognition and detection (multi-day offline training and real-time online testing to satisfy the technical requirements of cold-water coral group members during summer internship at Heriot-Watt University), code optimization and improvement to build GPU computation for processing huge image datasets and edge enhancement for feature-based maps, finally intensive nature analysis for different coral classes in variant aquatic environments (deep study on failure classification results based on physical properties and environmental correlation of target corals). 
\appendix
\chapter{Datasets}

\section{Moorea Labeled Corals}

\subsection{Coral classes}
\begin{figure}[h!]
\centering
	\includegraphics[width=0.75\textwidth]{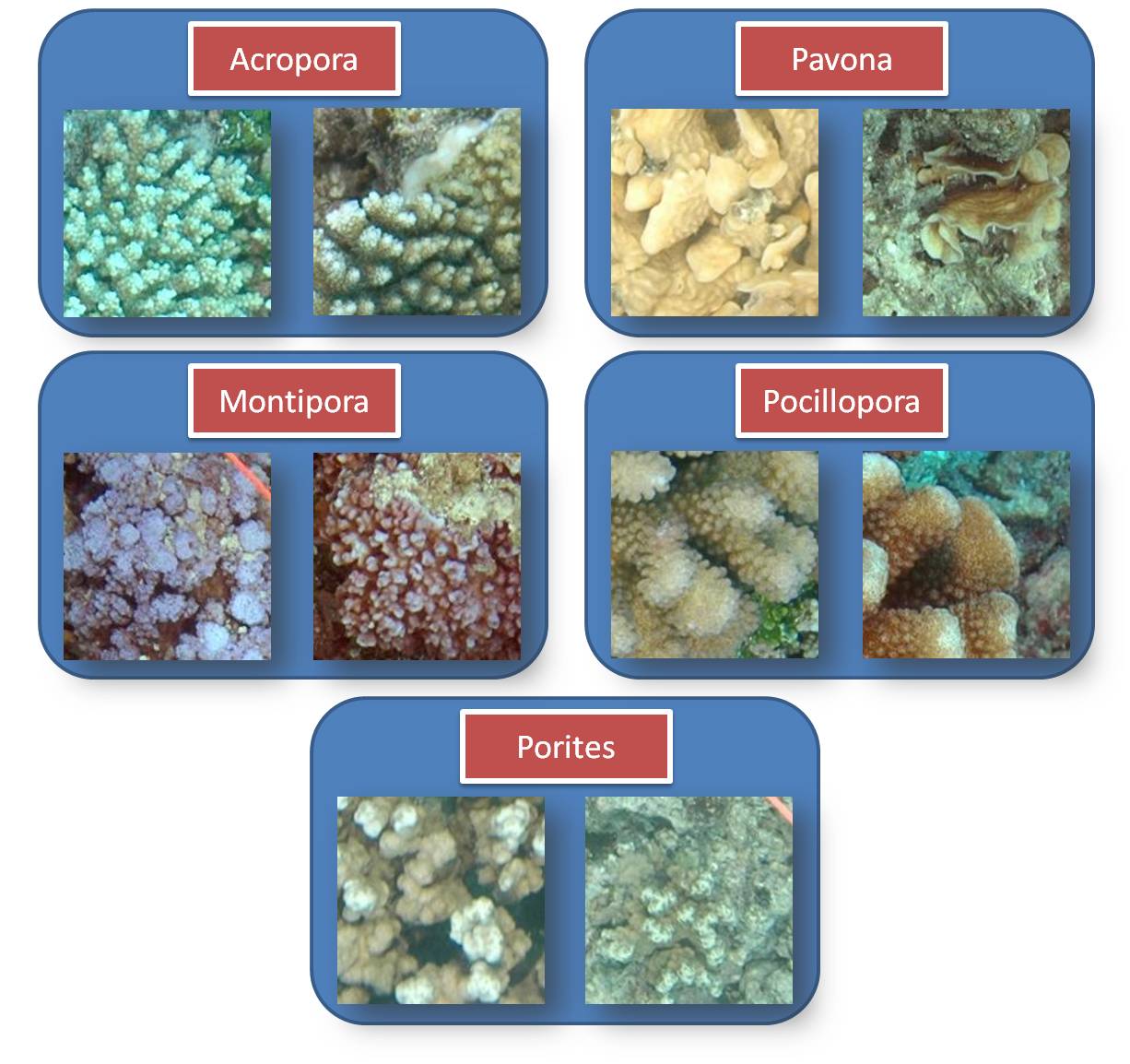}
\end{figure}

\newpage

\subsection{Non-coral classes}
\begin{figure}[h!]
\centering
	\includegraphics[width=0.75\textwidth]{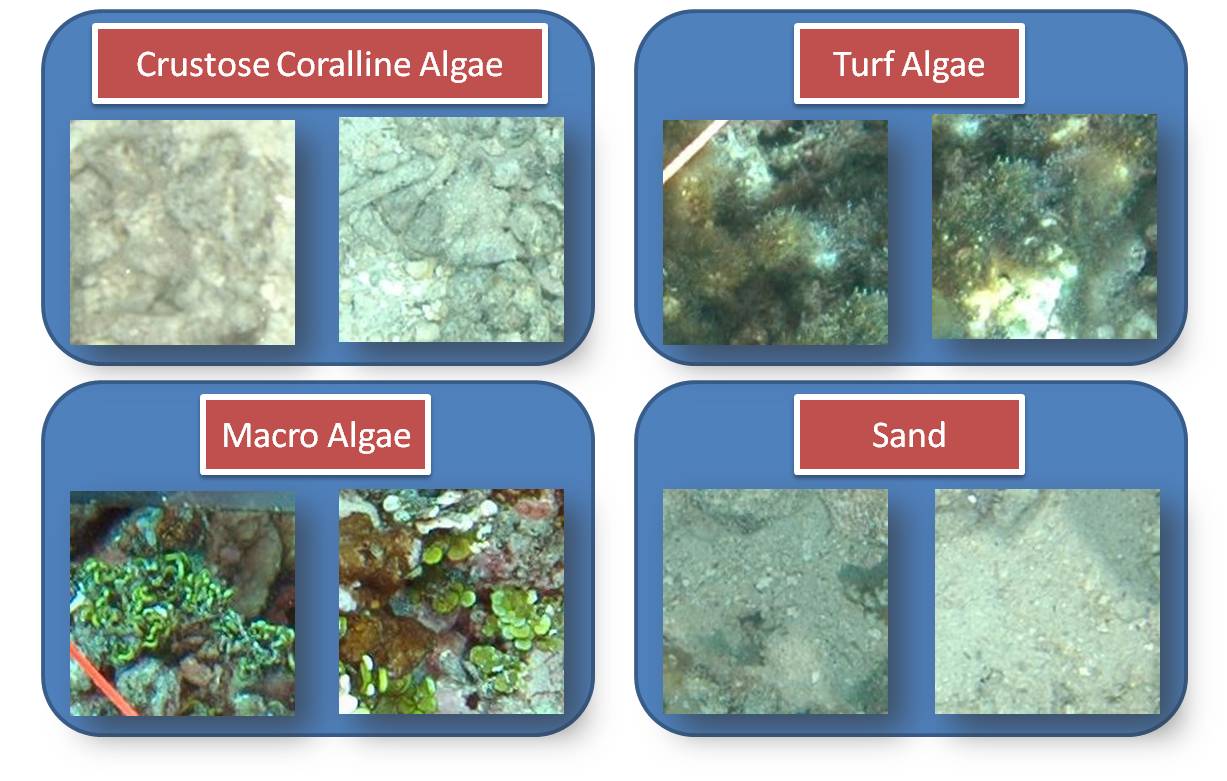}
\end{figure}

\newpage

\section{Atlantic Deep Sea}

\subsection{Coral classes}
\begin{figure}[h!]
\centering
	\includegraphics[width=0.75\textwidth]{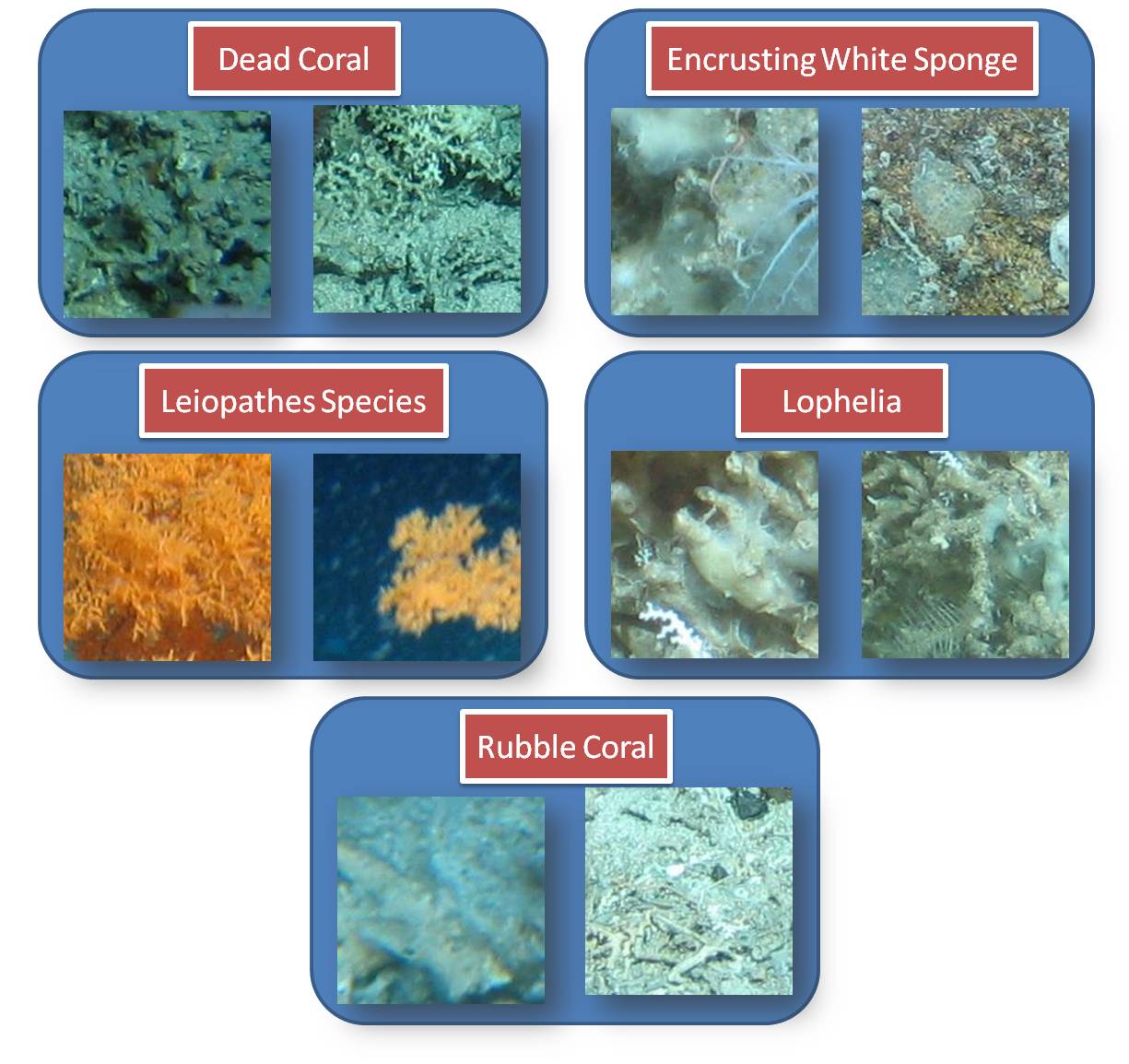}
\end{figure}
\newpage

\subsection{Non-coral classes}
\begin{figure}[h!]
\centering
	\includegraphics[width=0.75\textwidth]{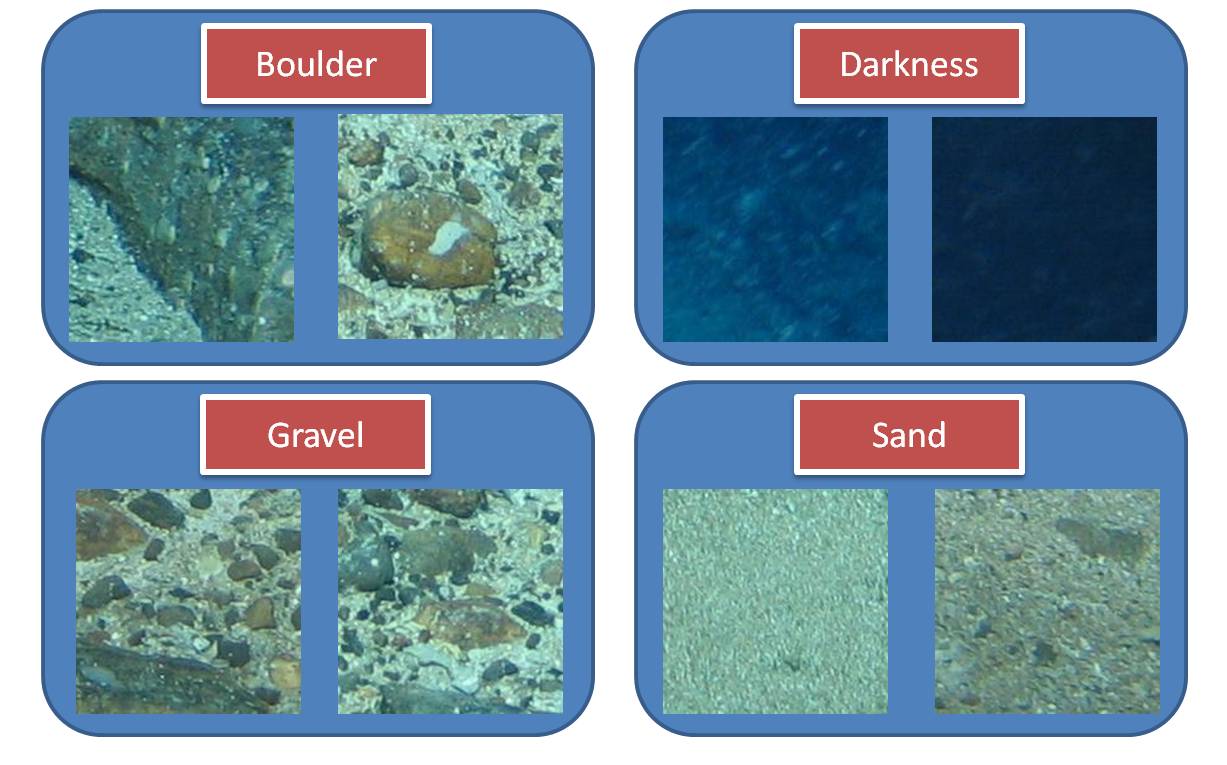}
\end{figure}

%   this is for BibTeX.  remove if you plan to write the references in the document
\bibliographystyle{ieeetr}
\bibliography{refs}

%adds the bibliography to the table of contents
\addcontentsline{toc}{chapter}
         {\protect\numberline{Bibliography\hspace{-96pt}}}

\end{document}